\documentclass[conference]{IEEEtran}
\IEEEoverridecommandlockouts
\usepackage{cite}
\usepackage{amsmath,amssymb,amsfonts}
\usepackage{algorithmic}
\usepackage{graphicx}
\usepackage{textcomp}
\usepackage{xcolor}
\def\BibTeX{{\rm B\kern-.05em{\sc i\kern-.025em b}\kern-.08em
    T\kern-.1667em\lower.7ex\hbox{E}\kern-.125emX}}
\usepackage{booktabs}
\usepackage[numbers,sort&compress]{natbib}
\usepackage{subfigure}
\usepackage{algorithm}
\usepackage{multirow, booktabs}
\usepackage{makecell}
\usepackage{calc}

\usepackage{color}
\usepackage[pdftex,linkcolor=blue,citecolor=blue,backref=page]{hyperref}
\usepackage{tabu}

\IEEEoverridecommandlockouts
\usepackage{threeparttable}

\begin{document}

\title{Pruning Filter via Gaussian Distribution Feature for Deep Neural Networks Acceleration}


 \author{

	\IEEEauthorblockN{Jianrong Xu$^{1,2}$, Boyu Diao$^{1}$\thanks{Boyu Diao and Bifeng Cui are Corresponding author of the article}
		, Bifeng Cui$^{2}$, Kang Yang$^{1}$, Chao Li$^{1}$, Hailong Hong$^{1}$}

	\IEEEauthorblockA{$^1$ Institute of Computing Technology, Chinese Academy of Sciences, Beijing, China}

	\IEEEauthorblockA{$^2$ Faculty of Information Technology, Beijing University of Technology, Beijing, China}

	\IEEEauthorblockA{\{jyfootprint, kang\_yang, duxingdaxia\}@foxmail.com,
		\{diaoboyu2012, lichao\}@ict.ac.cn, 	cbf@bjut.edu.cn
	}

}

\maketitle

\begin{abstract}
	Deep learning has achieved impressive results in many areas, but the deployment of edge intelligent devices is still very slow.
	To solve this problem, 
	we propose a novel compression and acceleration method based on data distribution characteristics for deep neural networks, 
	namely Pruning Filter via Gaussian Distribution Feature (PFGDF). 
	Compared with previous advanced pruning methods, PFGDF compresses the model by filters with insignificance in distribution, regardless of the contribution and sensitivity information of the convolution filter.
	PFGDF is significantly different from weight sparsification pruning because it does not require the special accelerated library to process the sparse weight matrix and introduces no more extra parameters. 
	The pruning process of PFGDF is automated. Furthermore, the model compressed by PFGDF can restore the same performance as the uncompressed model.
	We evaluate PFGDF through extensive experiments, on CIFAR-10,
	PFGDF compresses the convolution filter on VGG-16 by $ 66.62 \% $ with more than $ 90 \% $ parameter reduced, while the inference time is accelerated by $83.73\%$ on Huawei MATE 10.
\end{abstract}


\section{Introduction}
 In recent years, deep neural networks (DNN) have achieved remarkable performance in various fields, such as image classification
~\cite{simonyan2014very,wortsman2022model},
face recognition
~\cite{sun2018face,mare2021realistic},
	and object detection
	~\cite{deshmukh2018yolo,yang2021focal}, etc.
These jobs rely on deep networks with millions or even billions of parameters.
The availability of GPUs( Graphics Processing Units) with extremely high computing power plays a key role in their success.
These breakthroughs are closely related to the amount of existing training data and more powerful computing resources.
However, it is deployed in many deep learning applications (e.g. advertisement ranking, robots and self-driving cars).
DNN models are constrained by latency, energy and model size budget.
This is mainly because embedded devices or mobile devices cannot reproduce high computing power resources due to their size,
space and other constraints. This makes their computing and storage resources very valuable.
Fortunately, many methods~\cite{liu2021group, he2019filter,he2017channel}
have been successfully proposed to improve the hardware deployment of neural networks through the model compression.
Under the premise of ensuring that the performance of the model is not significantly reduced,
the model is compressed and accelerated by cutting the model's redundant parameter information or redundant structural information.

\begin{figure}[t]
	\begin{center}
		\includegraphics[width=1\linewidth]{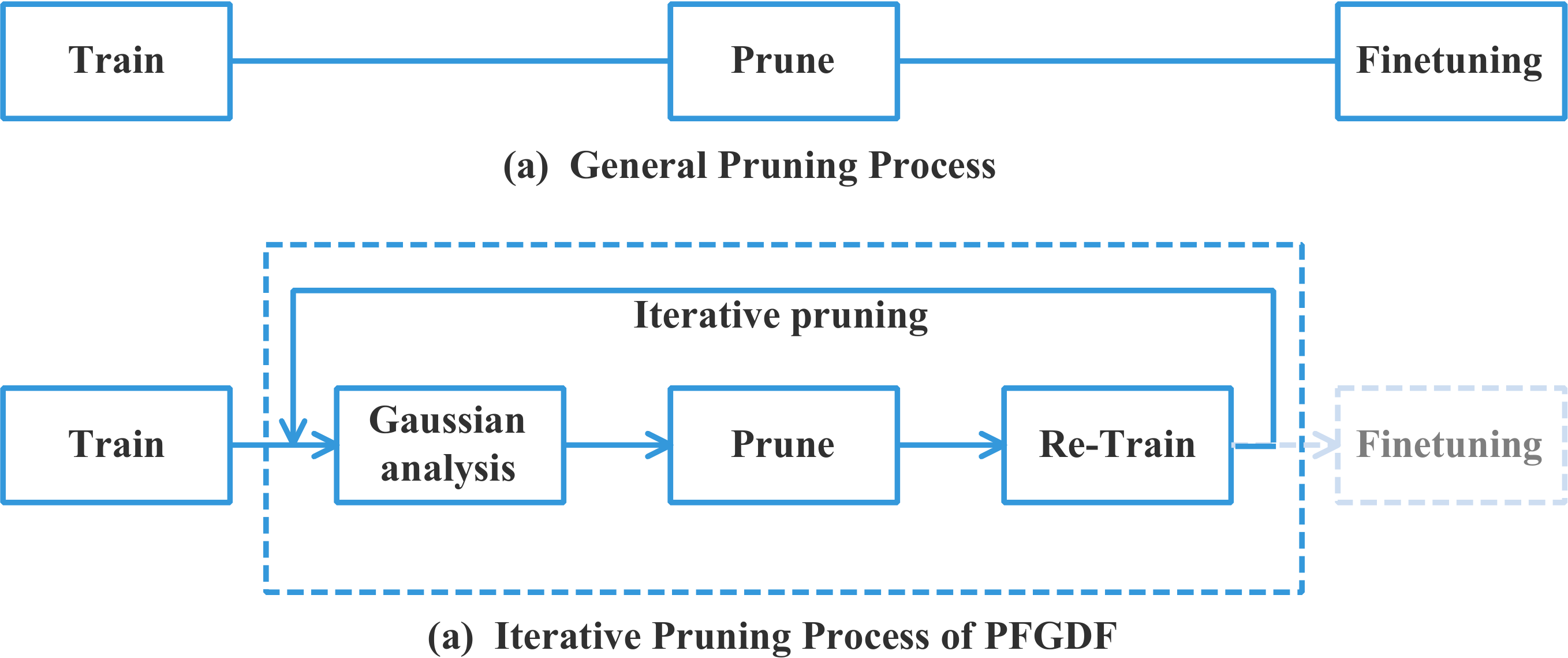}
	\end{center}
	\caption{
		Iterative Pruning Process of PFGDF VS General Pruning Process. After completing the pruning process of the trained model in a general iterative pruning process, fine-tuning is required but it is not necessary for PFGDF.}
	\label{fig:pruning procedure}
\end{figure}
In this work, we propose the Pruning Filter via Gaussian Distribution Feature (PFGDF) method.
This method focuses on trimming the model's convolution filter to reduce the model's parameter  and FLOPs.
Compared with unstructured sparse pruning~\cite{diffenderfer2021multi, sanh2020movement,zhu2017prune,han2015deep}, this method does not introduce sparseness,
so it does not require any special acceleration libraries, such as the Basic Linear Algebra Subprograms(BLAS) library,
nor do it require any special hardware to achieve the final compression and acceleration of the mode.
We find that the current advanced model compression methods are more focused on the aspects of
smaller-norm-less-important~\cite{he2018soft}, sensitivity~\cite{li2016pruning},
geometric distance~\cite{he2019filter},
but it lacks evaluation from the characteristic information of the learned model itself.
On the other hand, according to the knowledge we have at present,
some parameter distribution methods are adopted during the initialization stage of the model,
such as random initialization, Xavier initialization~\cite{glorot2018understanding},
MSRA initialization~\cite {he2018delving},
which can improve  the speed of model training convergence and the performance of the model network.
The converged model (like the initialization process) may also meet certain convergence feature distribution information.
Based on this consideration, we carry out related pruning and compression work.
We propose a model compression algorithm based on the distribution features of trained model,
named Pruning Filter via Gaussian Distribution Feature (PFGDF).
Different from the current methods, the process is based on features of filters,
the pruning process is automated without the need to manually set relevant compression hyper-parameter information.
Specifically, our method is a coupled pruning operation.
The trained model is analyzed layer-by-layer with distribution characteristics to ensure that the model after pruning can recover the performance of original model
as much as possible to cut out convolution filters that are not significantly distributed on this layer.
This means that the performance of the pruned model will not be lower than that of the original model before pruning, and fine-tuning is not necessary.
The implementation process is shown in Fig. ~\ref{fig:pruning procedure}.

Contributions. We have three contributions:

$ \bullet $ The trained model parameter information meets a certain statistical distribution rule. During the training process,
the distribution of weight information is always adjusted. After achieving fine inference performance at the end,
the distribution of parameter information almost obeyed a certain statistical distribution.

$ \bullet $ The method is automated and has no hyper-parameter information that needs to be manually adjusted.
The parameter were obtained by searching during the pruning process,
which do not need to be determined manually based on experience.
Compared with other methods~\cite{Liu_2017_ICCV, he2018soft, he2019filter,louizos2017learning}, we do not introduce super parameters before and after pruning. For example, regularized super parameters were introduced during training to conduct induction training, and pruning rate super parameters should be set during pruning. The proposed method is to extract the statistical performance characteristics of the trained model before pruning and performs the model compression,
so that it is easier to implement.

$ \bullet $ After pruning the entire  model, fine-tuning is not necessary. Even without fine-tuning, the pruned model still has similar reasoning performance.
If the model is fine-tuned, the accuracy of the final model might still improve.
\begin{figure}[t]
	\begin{center}
		\subfigure[initialization]{
			\begin{minipage}[t]{0.5\linewidth}
				\centering
				\includegraphics[width=1.5in]{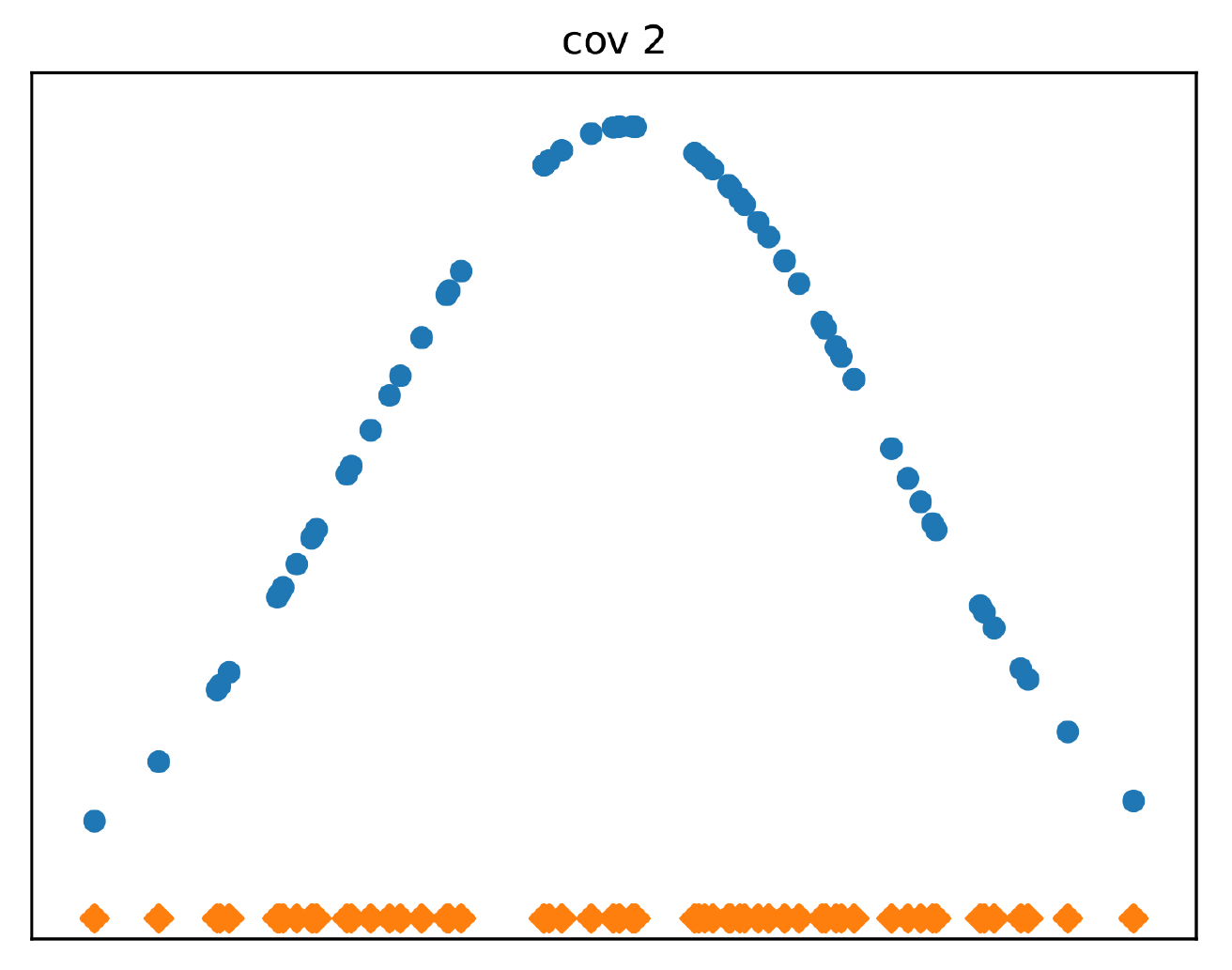}
			\end{minipage}%
		}%
		\subfigure[16 epochs]{
			\begin{minipage}[t]{0.5\linewidth}
				\centering
				\includegraphics[width=1.5in]{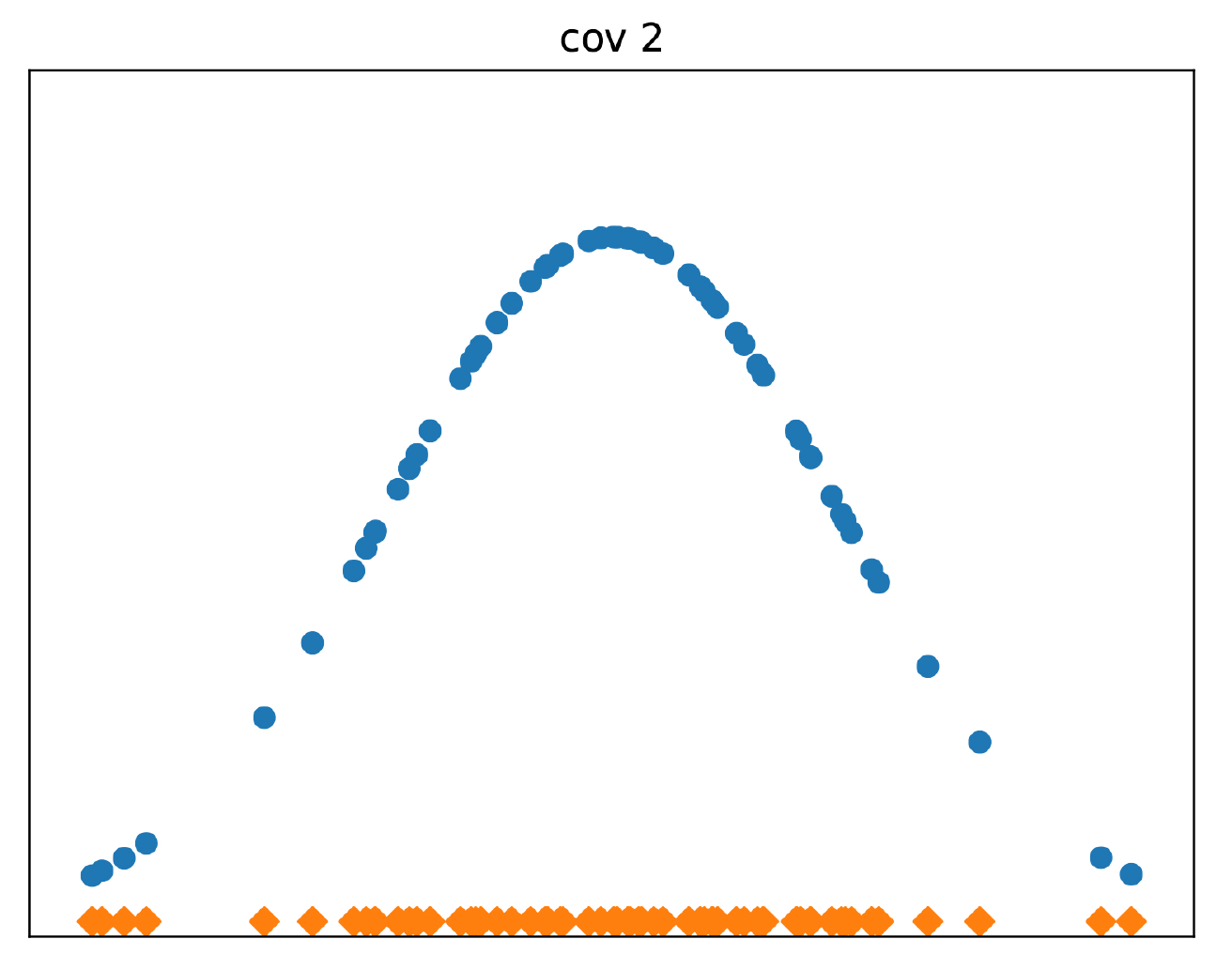}
			\end{minipage}%
		}%

		\subfigure[32 epochs]{
			\begin{minipage}[t]{0.5\linewidth}
				\centering
				\includegraphics[width=1.5in]{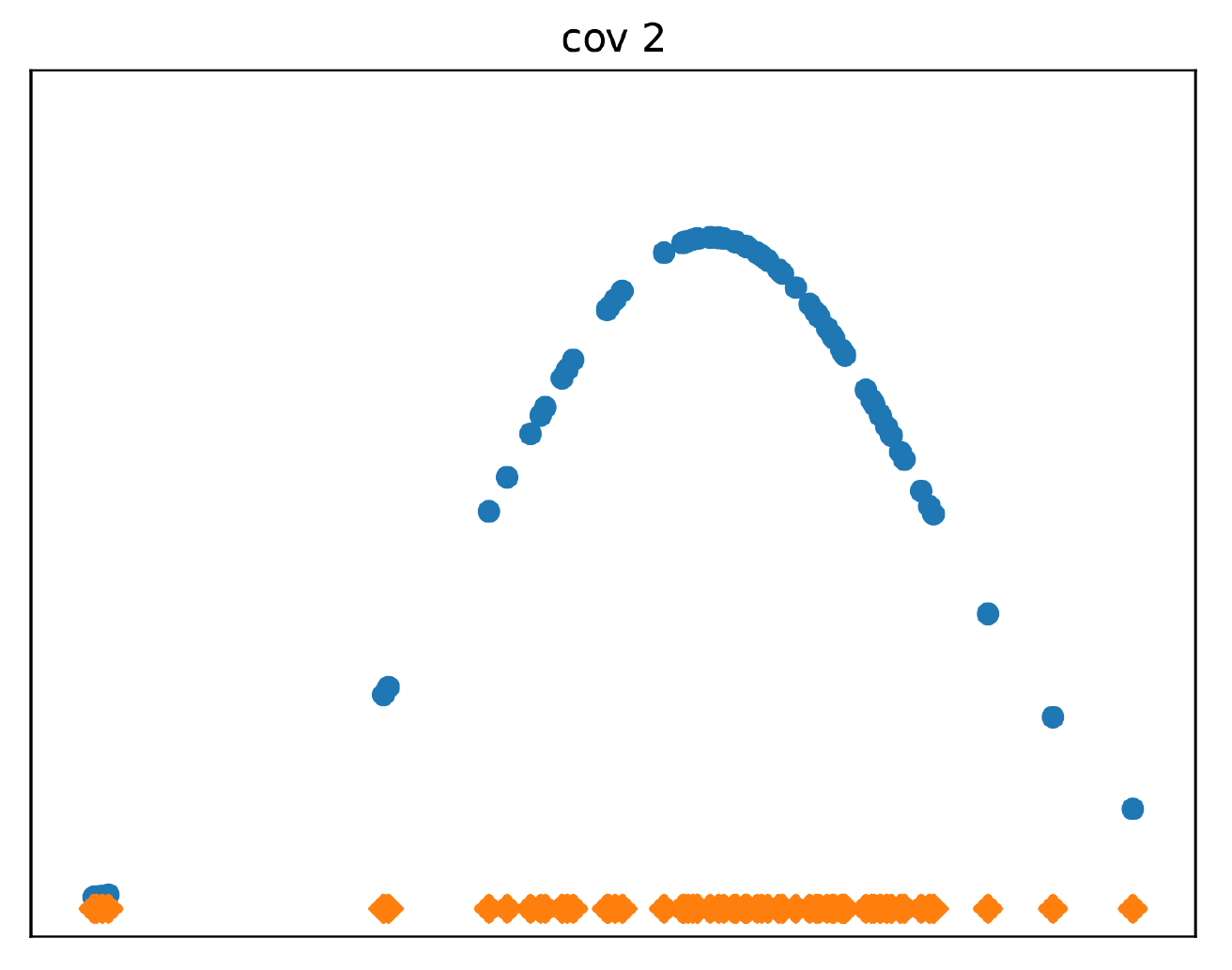}
			\end{minipage}
		}%
		\subfigure[160 epochs]{
			\begin{minipage}[t]{0.5\linewidth}
				\centering
				\includegraphics[width=1.5in]{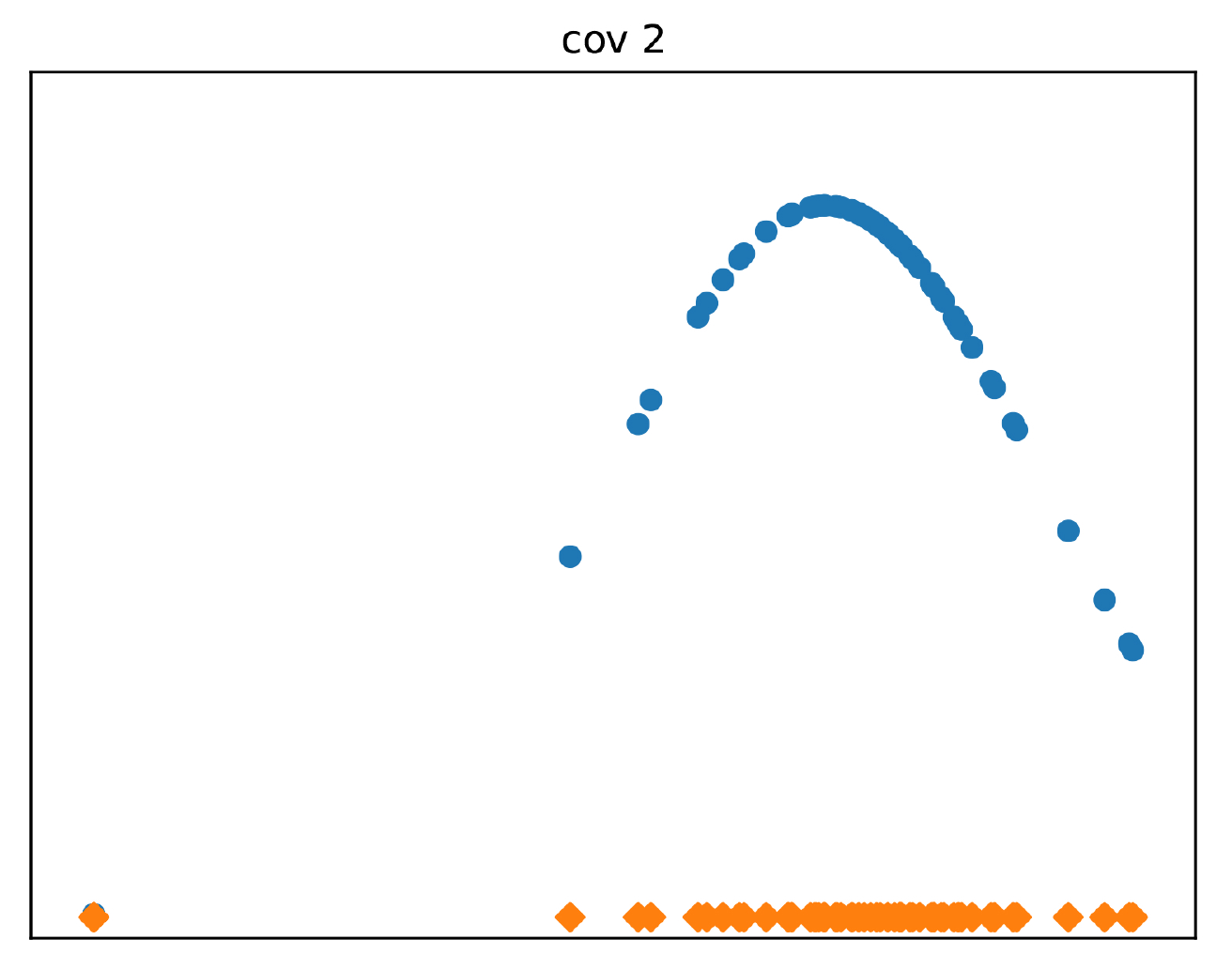}
			\end{minipage}
		}%

	\end{center}
	\caption{
		Illustrating the convolution filter $ f (i, j) $ 's Gaussian distribution during training.
		In CIFAR-10, the distribution of the second-level convolution filter of VGG-16 during the training processes.
		Blue: Gaussian distribution characteristics changing.
		Orange: Density changing.
	}
	\label{fig:show_gaussian}
\end{figure}
%

\begin{figure*}
	\begin{center}
		\includegraphics[width=1\linewidth,height=0.44\linewidth]{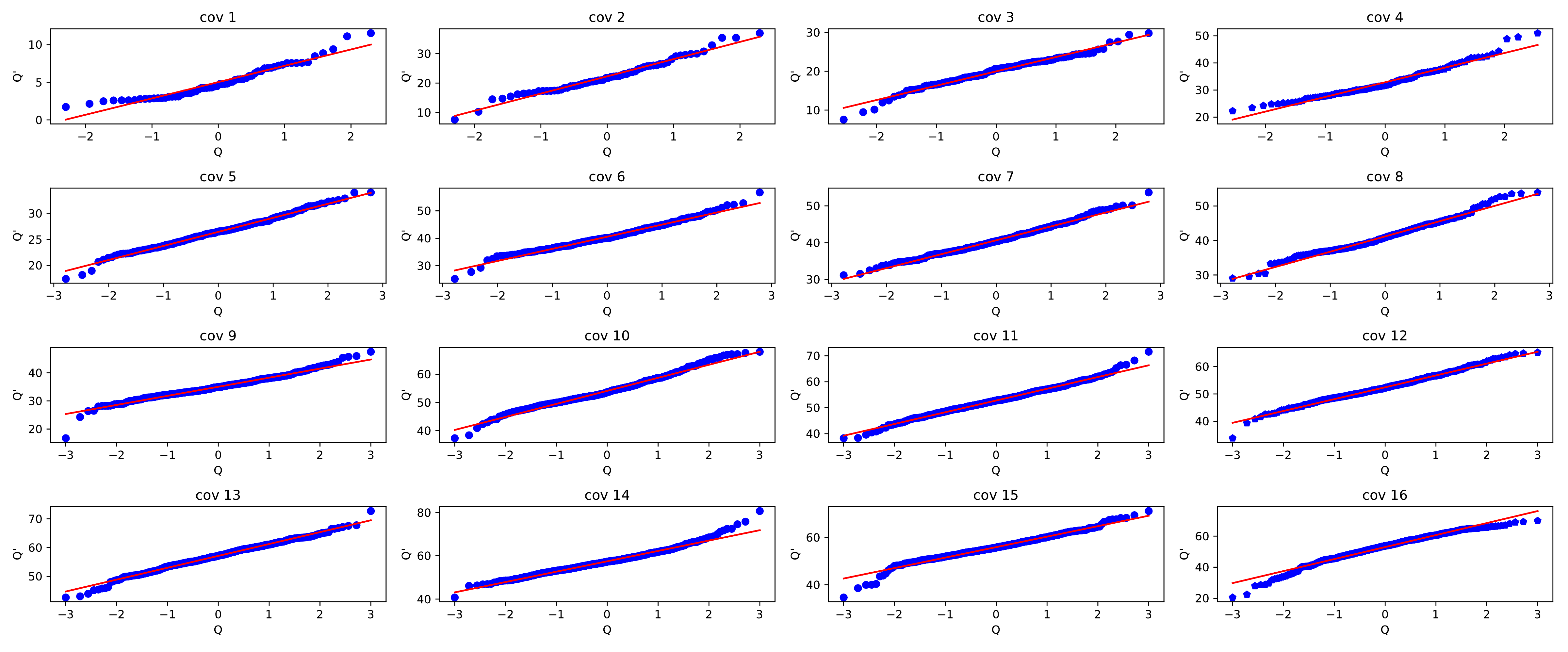}\\
	\end{center}
	\caption{
			Illustration of hypothetical rationality.
			On ImageNet LSVRC 2012 dataset, all the convolutional layers of VGG-19 are checked for the rationality of the Gaussian distribution on the QQ diagram.
			The points corresponding to $ Q-Q'$ are approximately distributed near the straight line.
	}
	\label{fig:vgg_gaussian_check}
\end{figure*}

\begin{algorithm}[tb]
	\caption{A algorithm for minimizing Eq.~\ref{eq:argmin_Q(x)_new}.}
	\label{alg:Framwork}
	\textbf{Input}: Training set {$ \left\{(x_{i}, y_{i}) \right\}$}\\
	\textbf{Parameter}:auto-prune parameter~$\alpha$\\
	\textbf{Output}: The compact model
	\begin{algorithmic}[1] 
		\STATE Compute layers L
		\STATE $\alpha \gets default$
		\STATE $prune\_layer \gets 0$
		\WHILE { $each\_layer\_index \in L$}

		\FOR {$prune\_layer$ $\in$ $T$}
		\STATE Compute $\mu, \sigma$
		\STATE Gauss analysis
		\STATE Find $h(j)$ filters that satisfy~\ref{eq:argmin_Q(x)_new}
		%
		\ENDFOR
		\STATE Re-train
		\IF{$pruned\_acc\geqslant original\_acc$}
		\STATE $each\_layer\_index \gets +1 $
		\STATE $\alpha \gets default$
		\ELSE
		\STATE Adjust $\alpha$
		\ENDIF
		\ENDWHILE
	\end{algorithmic}
\end{algorithm}

\section{Related work}

\textbf{Network Pruning}.
It is mainly used to solve the complexity of the model, the cost of calculation of the model
and to some extent can reduce the over-fitting issue of the model.
The process of network pruning can be divided into three steps: Train, Prune, and Fine-tuning.
Currently in the field of pruning, related methods can be roughly divided into two categories: unstructured pruning and structured pruning.
Unstructured pruning was first proposed by ~\cite{hanson1989comparing} for the magnitude-based pruning method, which applies weight decay
associated with its absolute value to each hidden unit in the network to minimize the number of hidden units. After that,
~\cite{lecun1990optimal} and ~\cite{hassibi1993second} respectively proposed OBD and OBS methods.
They evaluate the importance of weights in the network  based on the second derivative of the loss function relative to the weight (the Hessian matrix for the weight vector) and then prune it.
~\cite{han2015learning} proposes to prune network weights with small magnitude, and the technique is further incorporated  into
the "Deep Compression" pipeline (~\cite{han2015deep}) to obtain highly compressed models.
~\cite{louizos2017learning} proposes to encourage the model to become more sparse in the training process by introducing L0 regular terms to the objective function and designing the HARD CONCRETE DISTRIBUTION method during model training.
~\cite{sehwag2020hydra}  considers the use of robust training and network pruning jointly, and proposes a method aware of the training and make the training target guide the pruning process.

In contrast, structured pruning methods prune at the level of channels or even layers.
~\cite{li2016pruning} uses the lasso regression method to guide the model weight sparseness,  tailor the sparse channel to achieve the compression of the model, and transform the pruning compression problem into an optimization problem.
~\cite{he2019filter} calculates the distance between a filter and the center of the filter to determine whether the filter  should be cropped. The method is a norm-independent filter evaluation method FPGM, which breaks the norm evaluation limitations.
~\cite{tang2021manifold} considers different input information and proposes a new paradigm method for dynamic pruning. The method embeds the manifold information of all instances in the pruned network space, so as to achieve efficient network compression.
However, compared with structured pruning, unstructured pruning compresses the model by changing the unimportant weight parameter of the model to zero and making the model sparse.It does not directly compress and accelerate the model  so that it requires a special  hardware/librarie ~\cite{han2016eie}.

\section{Method}
\subsection{Preliminaries}
Symbols and notations in this subsection are formally introduced.
Suppose a neural network has $ L $ layers.
$ N_ {i} $ and $ N_ {i + 1} (1 \leq i + 1 \leq L) $ are represented as the $ i_ {th} $ layer input and output channel information.
$ M_ {i} $ is expressed as the total number of convolution filters in the $ i_ {th} $ layer.
The height and width of the input feature map at level $ N_ {i} $ are represented as $ h_ {i} / w_ {i} $.
The convolutional layer of this layer converts the input feature map $ x_{i} \in \mathbb {R} ^ {N_{i} * h_{i} * w_{i}} $
into the output feature map $ x_{i +1} \in \mathbb {R} ^ {N_ {i + 1} * h_ {i + 1} * w_ {i + 1}} $.
The 3D convolution filters acting on the $ N_ {i} $ input channel are represented as $ \mathcal {F} _ {i, j} \in \mathbb {R} ^ {N_ {i} * k * k} $.
$ \mathcal {F} _ {i, j} $ is represented as the $ j ({1} \leq {j} \leq {M_ {i}}) $ convolution filter of the $ i $ layer.
Each convolution filter consistes of 2D convolution kernel $ \mathcal {K} \in {k * k} (e.g., 5 * 5) $.
The $ L1-norm $ of each convolution filter is expressed as
\begin{gather}
{f(i,j)=|\mathcal{F}_{i,j}|}
\end{gather}
The convolution filter of this layer can be expressed as
\begin{gather}
{\mathcal{F}_{i} = \left\lbrace{ f(i,j) }\right\rbrace }
\end{gather}

\subsection{Analysis and Hypothesis}
We find in the course of the experiment that the trained baseline model, the $ L1 $ norm $ f(i, j) $
of any layer of the convolution filter of $ N_{i} $, meets certain statistical characteristics in the distribution.
We make $ {f (i, j)} \in {x} $, take $ Q (x) $ as a Gaussian function, then we have
\begin{gather}
{Q(x)=\frac{1}{\sigma~\sqrt{2~\pi}}e^{-\frac{(x-\mu)^2}{2\sigma^2 }} ,
	\quad x \sim \mathcal{N}\left(\mu, \sigma\right)}
\label{eq:Q(x)}
\end{gather}

Before and after training, the model of CNN(Convolutional Neural Network) presents the following characteristics on the Gaussian distribution function,
as shown in Fig.~\ref{fig:show_gaussian}:

1. With the model training, $ \mathcal {F} _ {i} $ appears sparsely distributed at both ends of the Gaussian function.

2. $\mathcal {F} _ {i} $ appears to converge to a certain center of the Gaussian distribution.

Based on such characteristics, we make the following hypothesis to guide the process of compression pruning :

\textbf {Hypothesis 1: Distribution Hypothesis}.
$\mathcal{F}_{i} $ of any convolutional layer of the model obeys Gaussian distribution.

\textbf{Hypothesis 2: Convergence Hypothesis}.
Based on hypothesis 1,
we think that during training, $ \mathcal {F}_{i} $ should converge to a center of the Gaussian distribution,
and the convolution filters might be redundant if they can't converge to the center and could be compressed safely.
The convolution filter $ f (i, j) $ distributed near the center has better information extraction capability and contribution.
	 After pruning a layer,  $ f (i, j) $ distributed near the center can be relearned to obtain the information of convolution filters  that are considered to be redundant  but in fact they are useful convolution filter  information, so as to achieve model performance recovery.

Based on the above hypothesis theory,
we think that there always exists an interval $ [a, b] $ containing important information in the convolution layer,
and the convolution filter $ f (i, j) $ outside this interval is the filter that could be pruned,
and the wrongly cropped convolution filter can be relearned by the filter in the interval $ [a, b] $ during performance recovery phase.

\textbf {Rationality of the distribution hypothesis}.
We perform a rough verification of the convolution filter distribution of each layer.
In this paper, we show the rationality of the convolutional layers of VGG-19 on the Gaussian distribution,
as shown in Fig.~\ref {fig:vgg_gaussian_check}.
We find that the convolution filters of these layers are approximately distributed near a straight line on the QQ diagram. A few convolution filters have
large deviations at both ends of the line.
But in the overall data, they are almost distributed at both ends of the line.

Through Fig.~\ref{fig:show_gaussian} and Fig.~\ref{fig:vgg_gaussian_check} experimental phenomenon and the above analysis, we think that the hypotheses are reasonable and accept it.
In our paper, we pay more attention to the relationship between these hypotheses and model compression.
It can be used to guide the process of compression pruning, and achieve the purpose of model optimization.

\begin{figure}[t]
	\centering
	\includegraphics[width=0.9\linewidth]{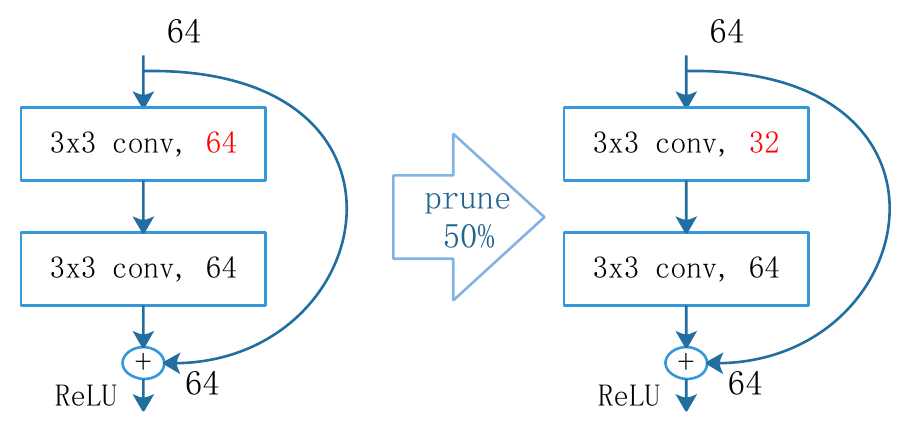}\\
	\caption{  Illustration of the ResNet pruning strategy.
		For each residual block, we do not prune the convolutional layer of shortcut connection partion, keeping the block output dimension unchanged}
	\label{fig:prune_resnet}
\end{figure}

	\subsection{Prune Framework}
	Framework of Pruning of PFGDF.
	Pruning is a classic approach used for reducing model complexity.
	Although there are huge differences (such as different criteria in selecting what should be pruned), the overall framework is similar in pruning
	filters inside a deep neural network.
	It can be summed up in one sentence: evaluate the importance of each neuron according to the set criteria, delete those neurons that are considered redundant, and fine-tune the entire network.

\begin{table*}[t!]
	\caption{
		The accuracy and pruning effect of VGG-NET on CIFAR-10 and CIFAR-100.
		We calculate the filters, FLOPs and parameters of PFGDF for VGG-NET.
		Ours A is the result without fine-tuning training, and Ours B is
		the result of fine-tuning from ous A.
	}
	\vspace*{-1.0mm}
	\centering
	\noindent\resizebox{1.0\linewidth}{!}{
		\tabulinesep=0.8mm
		\renewcommand{\arraystretch}{1.4}
		\begin{tabu}{lllrrrrrrrr}
			\Xhline{0.1em}
			\textbf{Dataset}           & \textbf{Model}          & \textbf{Result} & \textbf{Accuracy} & \textbf{Filters} & \textbf{Pruned} & \textbf{Parameters} & \textbf{Pruned} & \textbf{FLOPs} & \textbf{Pruned} \\
			\Xhline{0.1em}
			\multirow{6}{*}{CIFAR-10}  & \multirow{3}{*}{VGG-16} & Base            & 93.25\%           & 4224             & -               & 1.50E+07            & -               & 1.88E+09       & -               \\
			&                         & ours A          & 93.48\%           & 1410             & 66.62\%         & 1.38E+06            & 90.81\%         & 5.60E+08       & 70.27\%         \\
			&                         & ours B          & 93.63\%           & -                & -               & -                   & -               & -              & -               \\
			\Xcline{2-10}{0.1em}
			& \multirow{3}{*}{VGG-19} & Base            & 93.45\%           & {5504}       & -               & 2.03E+07            & -               & 2.39E+09       & -               \\
			&                         & ours A          & 93.46\%           & {1878}       & 55.54\%         & 2.00E+06            & 90.14\%         & 6.51E+08       & 72.80\%         \\
			&                         & ours B          & 93.73\%           & -                & -               & -                   & -               & -              & -               \\
			\Xhline{0.1em}
			\multirow{6}{*}{CIFAR-100} & \multirow{3}{*}{VGG-16} & Base            & 72.83\%           & 4224             &   -            & 1.50E+07            &   -              & 1.88E+09       &    -             \\
			&                         & ours A          & 72.97\%           & 1535             & 62.38\%         & 1.88E+06            & 87.52\%         & 7.38E+08       & 59.82\%         \\
			&                         & ours B          & {72.99\%}        & -                & -               & -                   & -               & -              & -               \\
			\Xcline{2-10}{0.1em}
			& \multirow{3}{*}{VGG-19} & Base            & 71.28\%           & 5504             &   -              & 2.03E+07            &     -            & 2.39E+09       &     -            \\
			&                         & ours A          & 71.29             & 1535             & 72.11\%         & 1.94E+06            & 90.49\%         & 7.58E+08       & 68.34\%         \\
			&                         & ours B          & {71.54\%}        & -                & -               & -                   & -               & -              & -               \\
			\Xhline{0.1em}
		\end{tabu}
	}
	\vspace{.3em}
	\label{tab:vgg16_19}
	\begin{tablenotes}
		\item  {
			{Measured using GTX1080Ti GPUs}
		}
	\end{tablenotes}
\end{table*}

	But in PFGDF, iterative pruning is adopted. During the pruning process, it is always guaranteed that the performance of the model could be restored to the original level to simplify the model as much as possible.
	Therefore, fine-tuning is not necessary.
	In the next section, we will focus on the part of our dashed box in Fig.~\ref{fig:pruning procedure} and introduce the PFGDF method.
	Given a model needed to be pruned, it will be pruned layer by layer with analysis. We summarize our framework as follows:

	1. \textbf{Training the model needed to be pruned.} This step is to train a baseline model. In this step we will obtain a trained model and its accuracy information.
	The trained model will be used in the pruning process in step 2 to perform model compression.
	And its accuracy information is used to ensure that the model can restore the original performance during the compression process.

	2. \textbf{Iterative pruning.}

	2.1. \textbf {Gaussian analysis of the convolution filter}.
	Unlike the existing model pruning methods that are the artificial setting of the specified layer $ l $ or the overall pruning ratio $ \alpha $,
	we perform a Gaussian distribution analysis to the layer to $ {\mathcal {F} _ { i}} $ , and the results of the analysis  are utilized to guide our pruning of this layer.
	The key idea is that if we can adopt a subset of filters in layer $l$ to approximate the all filters in layer
	$l$, and the other filters out of this subset will be safely removed.

	2.2.\textbf{Pruning weak filters}.
	Weak filters in layer $l$'s and their corresponding channels in layer $l+1$ will be pruned away, resulting in a much smaller model compared to the original model.
	The structure of the model after pruning is different from the original before pruning, and has  fewer filters and channels.
	New 3D-filter matrices  are created for both the $l$ and $l+1$ layer,
	the remaining filter weights are copied to the new model, and the weights will be reinitialized to restore model performance.

	2.3.\textbf{Retraining the pruned layer}.
	Fine-tuning is a necessary operation to restore the model's performance which is destroyed during pruning operations.
	However, ~\cite{DBLP:journals/corr/abs-1803-03635}
	and \cite{liu2018rethinking} recent research results indicate that after each pruning,
	the weight of the remaining nodes is reset to the initialization weight, which can increase the model weight probability of winning.
	Therefore, retraining operation is utilized so as to better allow the subset of convolution filters in the $ l $ layer to recover the original ability
	,and in order to remedy for the part of the convolution filters that are useful but pruned away.

	2.4.\textbf{Iterate to step 2.1 to prune the next layer until finishing all layer}

	3.\textbf{Fine-tuning for higher performance}
	It's worth noting that this step is not necessary.
	Because after the pruning in step 2.4, the pruned model has been able to recover to the same performance as the original model without pruning.
	The fine-tuning operation is based on the pruning model generated in step 2.4 for further weight parameter adjustment and learning update.
	This operation hopes that the model after pruning has better reasoning ability.

\subsection {Convolution filter selection}
Based on the previous analysis, the convolution $ {\mathcal {F} _ {i}} $ of any layer of the trained baseline model is derived from the distribution $ {Q (x)} $.
The distribution learned by the model at the $ i $ layer is $ Q_{i} (x) $, then ~\eqref{eq:Q(x)} at this layer means that the equivalent is:
\begin{gather}
{{Q_{i}(x)}, \quad {h(j)}\in{x},{{x}\in{T}} }
\end{gather}
$ T $ is the set of the information learned by the $ i_ {th} $ layer convolution filter.
Based on hypothesis theory, we implement the cropping of the model, which is to find a subset
${{x}\in{S}}, S \subseteq T=\left\{t_{1}, t_{2}, \cdots, t_{M_{i}}\right\}$.
Then we do not need any $h(j)$ if $h(j) \notin S$ and these filters will be safely removed
without damaging  the performance of the model.
The problem become the following optimization problem:
\begin{gather}
\arg \min \left\{Q_{i}(x), h(j) \in S\right\}
\label{eq:argmin_Q(x)}
\end{gather}
If $ h (j) $ is not close to a certain center of the gaussian distribution, it might be redundant.
Therefore, the search for the subset $ S $ is equivalent to deleting the redundant data on the left and right ends of the distribution $ T $, which is expressed as:
\begin{gather}
\begin{array}{l}{\mathrm{S}=\mathrm{T}-\mathrm{T}_{\mathrm{left}}-\mathrm{T}_{\mathrm{right}}} \\
{\mathrm{T}_{\mathrm{left}}=\left\{\mathrm{t}_{1}, \mathrm{t}_{2}, \mathrm{t}_{3}, \ldots, \mathrm{t}_{\mathrm{m}}\right\}} \\
{\mathrm{T}_{\mathrm{right}}  =\left\{\mathrm{t}_{\mathrm{p}+1}, \mathrm{t}_{\mathrm{p}+2}, \mathrm{t}_{\mathrm{p}+3}, \ldots, \mathrm{t}_{\mathrm{M_{i}}}\right\}}\end{array}
\label{eq:S}
\end{gather}

Here,
$\mathrm{T}=\mathrm{S} \cup \mathrm{T}_{1 \mathrm{eft}} \cup \mathrm{T}_{\mathrm{right}}$
,
and
$\mathrm{S} \cap \mathrm{T}_{1 \mathrm{eft}} \cap \mathrm{T}_{\mathrm{right}}=\varnothing$.
It is not easy for us to find $\mathrm{T}_{\mathrm{left}}$ and $\mathrm{T}_{\mathrm{right}}$.
We express ~\eqref{eq:S} equivalently as:

\begin{gather}
\begin{array}{l}{
	\mathrm{S}=\mathrm{T}_{\text {middlle }}}
\\ {\mathrm{T}_{\text {middlle }}=\left\{\mathrm{t}_{\mathrm{m}+1}, \mathrm{t}_{\mathrm{m}+2}, \mathrm{t}_{\mathrm{m}+3}, \ldots, \mathrm{t}_{\mathrm{p}}\right\}}\end{array}
\end{gather}

According to the hypothesis, $ \mathrm {T} _ {\text {middlle}} $ is distributed near
a certain center of the gaussian distribution $ Q_ {i} (x) $.
For further simplification, we approximate that the center is gaussian mean $ \mu $,
so $ \mathrm {T} _ {\text {middlle}} $ belongs to the bilateral symmetrical interval of $ \mu $.
Therefore, equation~\eqref{eq:argmin_Q(x)} is simplified to:
\begin{gather}
\arg \min \left\{Q_{i}(x), h(j) \in \left({\mu-{\alpha}{\sigma}}, \mu + {\alpha}{\sigma}\right)
\right\}
\label{eq:argmin_Q(x)_new}
\end{gather}

It should be noted that the $ h(j) $ of each layer is distributed in $ {\left (\mu - {\alpha}{\sigma}, \mu + {\alpha} {\sigma} \right)} $. The numbers on both sides of the interval are not symmetrical.

	\subsection {Pruning strategy}
	There are two main types of deep neural network structures: traditional convolutional network structures and the latest structural variants(shortcut block).
	In the former, the final analysis results are output through FC(Fully Connected Layer), such as the networks AlexNet ~\cite{krizhevsky2012imagenet},
	VGG-Net~\cite{simonyan2014very}.
	The latter such as ResNet~\cite{he2016deep},
	GoogleNet ~\cite{szegedy2015going} and other advanced network structures,
	adopt the global average pooling layer instead of FC ~\cite{lin2013network},
	and some novel network blocks are used in the network.
	For the above two different types of networks, we adopt different pruning methods.
	For the traditional convolutional network structure, we prune it directly.
	For the latest structural variants, due to the existence of some special structural connections in its network structure,
	there are certain restrictions on pruning of PFGDF. For example, the combination between shortcuts needs to maintain the consistency of the number of channels
	(see ~\cite {he2016deep} for more details).
	In order to keep the model consistent with the original, we do not prune them but kept some information about these structures. The details of this part are shown in Fig.~\ref{fig:prune_resnet}.

\begin{table*}[t!]
	\caption{
		The accuracy and pruning effect of ResNet are showed in CIFAR-10.
		We calculate the filters, FLOPs and parameters for ResNet of different depths.
		Ours A is the result of no fine-tuning training, and Ours B is fine-tuning result.
	}
	\centering
	\noindent\resizebox{1.0\linewidth}{!}{
		\tabulinesep=0.8mm
		\renewcommand{\arraystretch}{1.28}
		\begin{tabu}{lllrrrrrrr}
			\Xhline{0.1em}
			\textbf{Dataset}    &\textbf{Model}           & \textbf{Result} & \textbf{Accuracy} & \textbf{Filters} & \textbf{Pruned} & \textbf{Parameters} & \textbf{Pruned} & \textbf{FLOPs} & \textbf{Pruned} \\
			\Xhline{0.1em}
			\multirow{6}{*}{CIFAR-10} & \multirow{3}{*}{ResNet-20} & Base            & 91.79\%           & 688              & -               & 2.70E+05            & -               & 2.45E+08       & -               \\
			& & ours A          & 91.85\%           & 634              & 7.85\%          & 2.27E+05            & 16.00\%         & 2.04E+08       & 16.62\%         \\
			& & ours B          & 91.91\%           & -                & -               & -                   & -               & -              & -               \\
			\Xcline{2-10}{0.1em}
			&\multirow{3}{*}{ResNet-32} & Base            & 91.97\%           & 1136             & -               & 2.70E+05            & -               & 2.45E+08       & -               \\
			& & ours A          & 91.99\%           & 887              & 21.92\%         & 1.22E+05            & 54.64\%         & 1.41E+08       & 42.32\%         \\
			& & ours B          & 92.01\%           & -                & -               & -                   & -               & -              & -               \\
			\Xhline{0.1em}
		\end{tabu}
	}
	\vspace{.3em}
	\label{tab:resultOnResNet}
\end{table*}

\begin{table}[t!]
	\caption{
		{Compare VGG network with other method on CIFAR dataset.}
	}
	\centering
	\noindent\resizebox{1\linewidth}{!}{
		\tabulinesep=0.8mm
		\renewcommand{\arraystretch}{1.4}
		\begin{tabu}{lllrrr}
			\Xhline{0.1em}
			\textbf{Dataset}           & \textbf{Model}           & \textbf{Method}       & \textbf{Top-1 Acc.} & \textbf{\#Param $\downarrow$} & \textbf{\#FLOPs $\downarrow$} \\
			\Xhline{0.1em}
			\multirow{13}{*}{CIFAR-10} & \multirow{9}{*}{VGG-16}
			& WooJeong $et \ al$,2020~\cite{kim2020neuron}         & -0.54\%    & 63.7\%          & - \\
			&                          & Zhao $et \ al$,2019~\cite{zhao2019variational} & -0.07\%    & 73.34\%          & 39.10\% \\
			&                          & GAL-0.05,2019~\cite{lin2019towards}            & +0.19\%    & 77.60\%          & 39.60\% \\
			&                          &Huang  $et \ al$,2018\cite{huang2018data}       & +0.33\%    & 66.70\%          & 36.30\% \\
			&                          &Xu  $et \ al$,2018~\cite{xu2018globally}        & +0.10\%    & -                & 61.45\% \\
			&                          & \textbf{our A}                  & \textbf{+0.23\%}    & \textbf{90.81\%} & \textbf{70.27\%} \\
			&                          & \textbf{our B}                  & \textbf{+0.38\%}    & \textbf{-} & \textbf{-} \\
			\Xcline{2-6}{0.1em}
			& \multirow{4}{*}{VGG-19}  & Liu $et \ al$,2017~\cite{Liu_2017_ICCV}        & -1.33\%    & 80.07\%           & 42.65\% \\
			&                          &  Wang $et \ al$,2019~\cite{wang2019eigendamage}& -0.19\%    & 78.18\%           & 37.13\% \\
			&                          & \textbf{our A}                           & \textbf{+0.01\%}    & \textbf{90.14\%} & \textbf{72.80\%} \\
			&                          & \textbf{our B}                           & \textbf{+0.28\%}    & \textbf{-} & \textbf{-} \\
			\Xhline{0.1em}
			\multirow{3}{*}{CIFAR-100} & \multirow{3}{*}{VGG-16}  & Zhao $et \ al$,2019~\cite{zhao2019variational} & +0.07\%    & 37.87\% & 18.05\% \\
			&                          & {Tao $et \ al$,2020~\cite{zhuang2020neuron}}    & {-0.42\%}    & {-} & {43.0\%} \\
			&                          & {WooJeong $et \ al$,2020~\cite{kim2020neuron}}    & {-1.67\%}    & {44.0\%} & {-} \\
			&                          & \textbf{our A}    & \textbf{+0.07\%}    & \textbf{87.52\%} & \textbf{59.84\%} \\
			&                          & \textbf{our B}    & \textbf{+0.16\%}    & \textbf{-} & \textbf{-} \\
			\Xhline{0.1em}
		\end{tabu}
	}
	\begin{tablenotes}
		\item  { {The higher the percentage, the better}}
	\end{tablenotes}

	\label{tab:method_vgg_compare}
\end{table}

\section{Experiment}

	\subsection{Experimental Settings}

	\textbf{Training setting}. On CIFAR-10, the experimental parameter setting is basically the same as
	the hyper-parameter setting of ~\cite{szegedy2016rethinking,li2016pruning}, and the experimental test is based on the
	open source model code $\footnote{https://github.com/bearpaw/pytorch-classification}$.
	In most cases, hyper-parameter settings in different models are the default settings, so they are basically the same.
	All networks are trained for 160 epochs and with batch size 64.
	 The learning rate is set to 0.1, and is multiplied by 0.1 at the 0.5 to 0.75 the total training epochs.
	Similarly, the configuration on the CIFAR-100
	 is almost the same.
	Meanwhile, all these networks are optimized by SGD(stochastic gradient descent) with weight decay $10^{-4}$ and momentum 0.9.

\textbf{Prune setting}.
In the filter pruning step,	PFGDF firstly loads the baseline model completed by training dataset in the training phase,
then performs statistical analysis of the data distribution layer by layer, and deletes unimportant convolution filters based on the results of the analysis.
Different with other methods~\cite{he2019filter,li2016pruning},
the pruning process is automatic with the default pruning parameters (from $ \alpha = 0.3 $)
which is confirmed using a grid algorithm without the need for human intervention.
The greedy algorithm strategy is adopted in the process of pruning, and ensuring that the accuracy of the original model can be restore.
It is worth noting that the cropping process started from the last layer of the most abstract convolutional layer of semantic information
and gradually operates towards the beginning of the network.
Each time the pruning completed the specified convolutional layer, we adopt the latest research results of ~\cite{DBLP:journals/corr/abs-1803-03635}
and ~\cite{liu2018rethinking} that is to re-initialize the convolutional layer after pruning,
and the other convolutional layers that have not been pruned will keep the original learned parameter, and then train the entire network with pruning process, it is worth noting that the pruning parameters(frome defulat $\alpha = 0.3 $ to the appropriate parameter for the $ l_{i} $ layer) are implemented based on training dataset, and each layer would get different parameters due to their different distribution characteristics.

	If the pruning granularity of the $ l_{i-1} $ layer is so large
that the performance of the model of the $ l_{i} $ layer will not be recovered,
PFGDF will automatically adjust the granularity of the last pruning operation in the $  l_{i-1} $ layer and prune again.

\subsection{Result on VGG}
We perform experiments on CIFAR-10 and CIFAR-100. Detail information {as shown in Table~\ref{tab:vgg16_19}}.
The baseline of VGG-16 achieves the classification accuracy $ 93.25 \% $ on CIFAR-10.
Without fine-tuning, the accuracy of PFGDF reaches $ 93.48 \% $, which is higher than the baseline accuracy.
$ 90.81 \% $ parameter information is compressed,
and $ 70.27 \% $ FLOPs is cropped. Compared with CIFAR-100,
the parameters and the compression of FLOPs on CIFAR-10 are larger.
After the pruning, there is no loss of accuracy.
The reason was that PFGDF, in the process of pruning, should ensure
that the model can restore the performance similar to the original model when pruning each layer.
Therefore, fine-tuning of the model after cutting is not necessary, and it will be directly deployed to the corresponding environment.
Ours A is the performance of the model after pruning, and ours B is a fine-tuning training from ours A.
It will be found that the effect of our B is almost better.
Compared to VGG-16, the cutting granularity of VGG-19 on FLOPs is larger, reaching CIFAR-10 and CIFAR-100 to $ 72.80 \% $ and $ 68.34 \% $ respectively.
In convolution filter compression, VGG-19 is only better than VGG-16 in CIFAR-100.

\subsection{Result on ResNet}
We also explore the effects of PFGDF on ResNet in Table~\ref{tab:resultOnResNet}.
We use ResNet-20 and ResNet-32 as our baseline models.
ResNet on the CIFAR-10 dataset have a total of 3 different residual blocks which are 32x32, 16x16, and 8x8 respectively.
Each residual block has the same number of convolution filters.
In experiment,
in order to ensure the integrity of the model structure that the residual block contains a lot of information,
PFGDF just prunes the first layer of each residual block ,
and do not take any action on the first convolution layer.

As shown in Fig.~\ref{fig:prune_resnet}.
On ResNet-20, the compression effect is not particularly obvious. The compression of parameters and FLOPs is higher than $ 16 \% $,
while the convolution filter only compresses $ 7.85 \% $.
After compression, the accuracy is better than the baseline.
In contrast, PFGDF has better compression on ResNet-32.
The parameter of $ 54.64 \% $ is pruned, and the FLOPs of $ 42.32 \% $ is compressed.

No matter it is on VGG or ResNet, the PFGDF always guaranteed that the test accuracy of the model is not lower than the original accuracy during pruning,
and a good achievement is achieved.

\subsection{Comparison with Other Method}

\begin{table}[t!]
	\caption{
		{Compare ResNet network with other method on CIFAR dataset.}
	}
	\vspace*{-1.0mm}
	\centering
	\noindent\resizebox{1\linewidth}{!}{
		\tabulinesep=0.8mm
		\renewcommand{\arraystretch}{1.4}
		\begin{tabu}{lllrrr}
			\Xhline{0.1em}
			\textbf{Dataset}  &	\textbf{Model}             & \textbf{Method} & \textbf{Top-1 Acc.} & \textbf{\#Param $\downarrow$} & \textbf{\#FLOPs $\downarrow$} \\
			\Xhline{0.1em}
			\multirow{7}{*}{CIFAR-10} & \multirow{4}{*}{ResNet-20} & Zhao $et \ al $,2019~\cite{zhao2019variational}    & -0.35\%   & 20.41\%  & 16.47\%                 \\
			&                         & He $et \ al $,2018~\cite{he2018soft}  & +0.04\%            & -                 & 15.20\%                  \\
			&                         & \textbf{our A}                  & \textbf{+0.06\%}   & \textbf{16.00\%}  & \textbf{16.62\%}          \\
			&                         & \textbf{our B}                  & \textbf{+0.01\%}   & \textbf{-}        & \textbf{-}               \\
			\Xcline{2-6}{0.1em}
			&\multirow{3}{*}{ResNet-32} & He $et \ al $,2018~\cite{he2018soft}     & +0.590\%            & -                   & 14.90\%                                    \\
			&                   & FPGM $et \ al $,2019~\cite{he2019filter}         & {-0.32\% }         & {-}    & {41.5\%}                          \\
			&                          & \textbf{our A}                    & \textbf{+0.02\% }    & \textbf{54.64\%}    & \textbf{42.32\%}                          \\
			&                          & \textbf{our B}                     & \textbf{+0.04\% }   & \textbf{-}          & \textbf{-}                                          \\
			\Xhline{0.1em}
		\end{tabu}
	}
	\vspace{.3em}
	\label{tab:experiment_resnetComparedWithOthers}
\end{table}

We compare our experimental results with other state-of-the-art model compression methods.
The results show that our experimental results have obvious advantages.
The accuracy of PFGDF during pruning(our A) always stays above the baseline. This means that there is no loss in performance of the pruned model.
After fine-tuning(our B), the accuracy of the model might still be further improved.
The search process for compressed hyper-parameter do not need to be determined by human experience,
and it is simpler to implement by using a grid search method to determine by a program.

As shown in Table~\ref{tab:method_vgg_compare}. For VGG-16 On CIFAR-10, in addition to WooJeong $et \ al$~\cite{kim2020neuron}, other compression methods need to make Fine-Tuning to achieve accuracy recovery after pruning the model, and the  recovery accuracy is still damaged. And for PFGDF is not necessary. The PFGDF has increased the accuracy of the model to $0.23 \%$ after the pruning, exceeding Zhao $et \ al$~\cite{zhao2019variational}, GAL-0.05~\cite{lin2019towards} and Xu  $et \ al$ ~\cite{xu2018globally}.
After fine-tuning on the PFGDF, the accuracy of the model is improved by $0.23 \%$, which is higher than that of WooJeong $et \ al$~\cite{kim2020neuron} and Huang  $et \ al$\cite{huang2018data}.In addition, most methods need to provide pruning hyper-parameter through experience, such as Zhao $et \ al$~\cite{zhao2019variational} and WooJeong $et \ al$~\cite{kim2020neuron}, while the pruning hyper-parameter in PFGDF is implemented by automation.
Furthermore, Zhao $et \ al$~\cite{zhao2019variational} uses the Variational CNN Pruning method to trim redundant channels.
Through the significant characteristics of the BN layer distribution,
which is similar to our method.
The difference is that our pruning method is based on the distribution characteristics of filters itself.
For VGG-16 on CIFAR-10, the amount of parameters is reduced by $ 73.34 \% $ and by $ 39.1 \% $ FLOP with Variational CNN Pruning, but the results of PFGDF are much better.
\begin{figure}[t!]
	\begin{center}
		\includegraphics[width=1\linewidth]{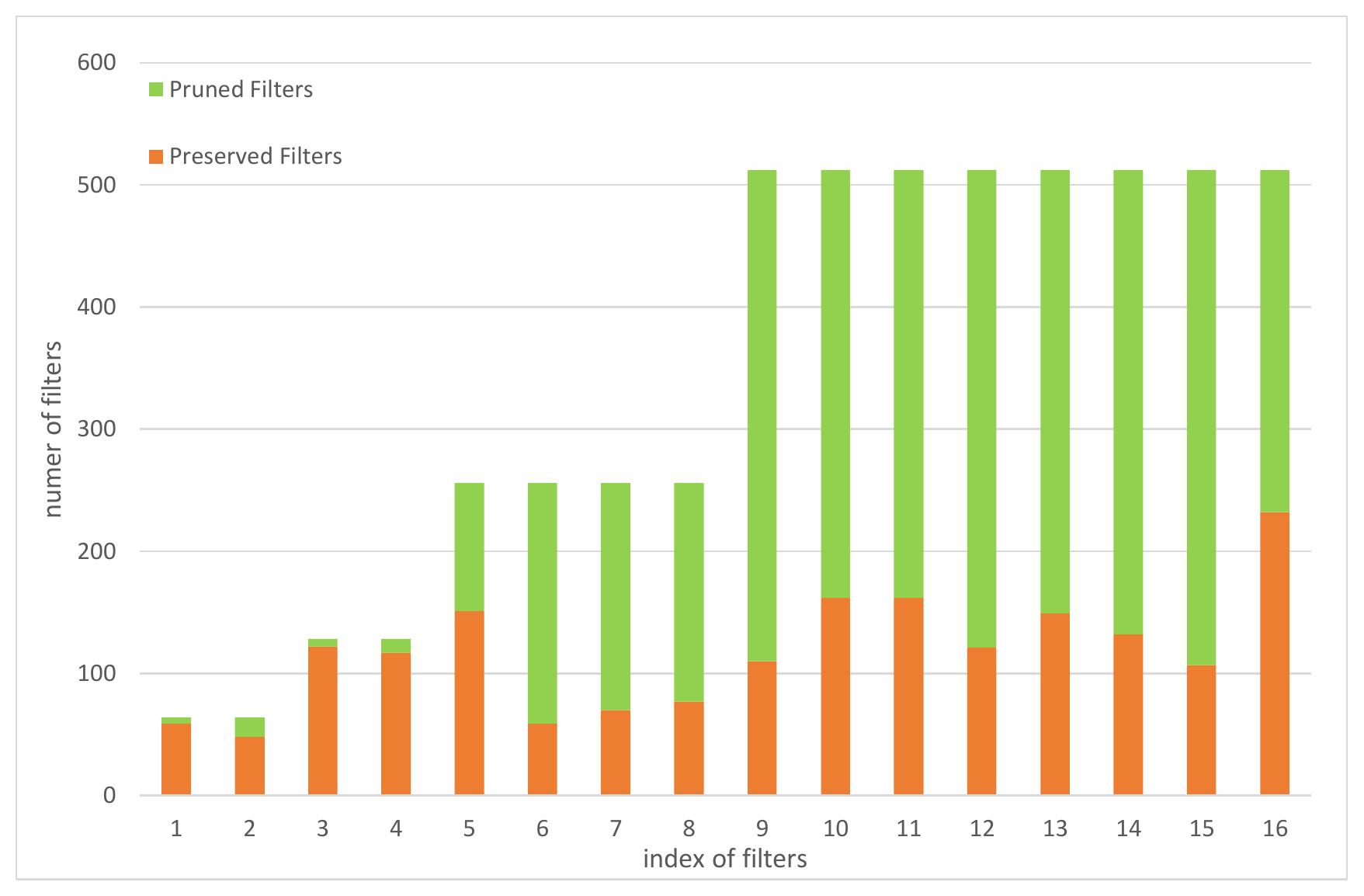}
	\end{center}
	\caption{VGG-19 model convolution filter pruning details on CIFAR-10}
	\label{fig:vgg19_pruning_details}
\end{figure}
As shown in Table~\ref{tab:experiment_resnetComparedWithOthers}.
We show the performance of ResNet on CIFAR-10.
Similarly, PFGDF on ResNet is better than the method of ~\cite{zhao2019variational}.
There is no loss of accuracy after pruning but in the fine-tuning stage,
the accuracy improved slightly by $ 0.01 \% $ on ResNet-20, which is still better than ~\cite{zhao2019variational}
and ~\cite{he2018soft}.
The compression effect of ResNet-32 is very significant, and the amount of parameters and FLOPs exceed $ 40 \% $.

These results indicate that PFGDF have significant compression acceleration for parameters and FLOPs.

\subsection{Result on neural style transfer}

	In this section, we will test whether the pruned model will affect the effect of neural network style transfer, and accelerated results will be reported here.
	We use the baseline model and pruning model of VGG on CIFAR-100 (see Table ~\ref{tab:vgg16_19} for details).
   Neural style transfer is an algorithm that combines the content of one image with the style of another image using CNN~\cite{NeuraArtisticStyle2015l}.
   We take two images, an input image and a content-image, and the algorithm will generate a target image that minimizes the content difference with the content image and the style difference with the style image.
   We select the feature map outputs by the 1st, 3rd, 5th, and 9th convolutional layers of VGG for loss calculation of style transfer.
	Durning forward propagation process, we input the content image and the target image to VGG which is trained by CIFAR-100 datasets. Then, the specified feature maps will be extracted to minimize the feature maps of the content image and its feature maps with  the mean-squared error to compute content loss.
	For style loss, as the same with content loss, the style image and target image are input into the VGG to generate a texture
	which match the style of the style image. The target image will be updated by minimizing the mean-squared error trough the Gram matrix which are between the style image and the target image.

\begin{figure*}
	\begin{center}
		\includegraphics[width=1\linewidth]{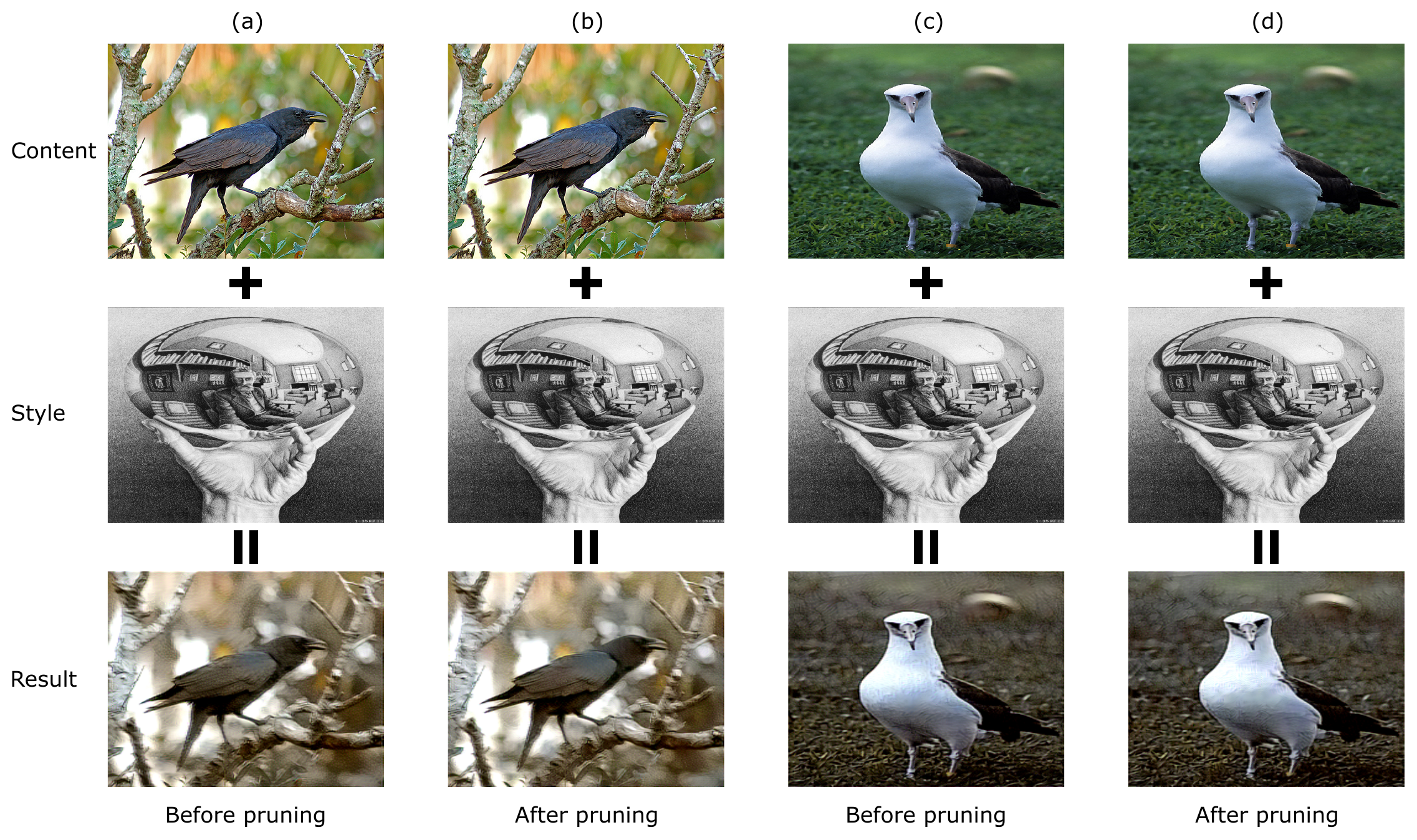}\\
	\end{center}
	\caption{
		Compare the effects of model before and after pruning on neural network style transfer.
		The model before and after pruning of VGG on CIFAR-100 dataset is used for neural network style transfer.
		 (a) and (b) are the results of VGG-19 model before and after pruning respectively.
		 (c) and (d)  are the results of VGG-16 model before and after pruning respectively.
	}
	\label{fig:neural_style_transfer}
\end{figure*}

	As shown in Fig.~\ref{fig:neural_style_transfer}. (a) and (c) are the results of neural network style transfer of the baseline model trained on CIFAR-100 by VGG-19 and VGG-16, respectively.
	Results of (b) and (d) are shown by neural network style transfer of the model of VGG-19 and VGG-16 pruning on CIFAR-100, respectively.
	Compared with the effect before and after pruning, the overall are generally similar although they have slight differences in effect.
	The network after pruning has fewer convolution filters than before pruning, meaning that the extracted information of the compressed model will be less when extracting abstract semantic information.
	In fact, when extracting features, the compressed model is still excellent in neural network style transfer.
	The experiment shows that if the compressed models can restore the same performance as the models before pruning,
	they still have the generalization ability similar to the models before compression.

\subsection{Pruning Analysis}

\begin{figure}[t]
	\begin{center}
		\includegraphics[width=1\linewidth]{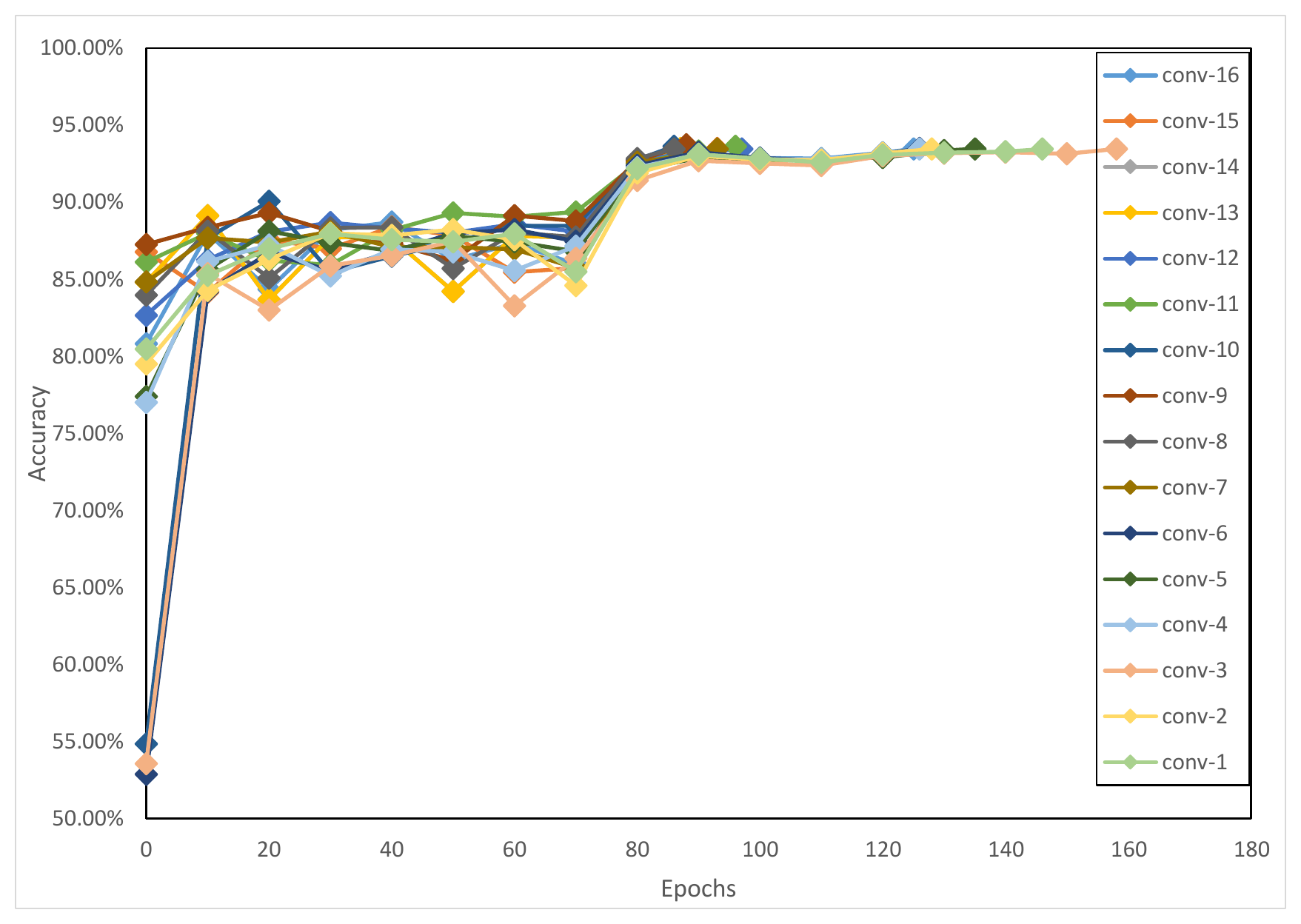}
	\end{center}
	\caption{
		Accuracy recovery curves of the compression. We reveal the details of
		the accuracy recovery of the model during the pruning compression of VGG-19 on CIFAR-10.
	}
	\label{fig:experiment_vgg19_retraining_acc}
\end{figure}
%
%


%
As shown in Fig.~\ref{fig:vgg19_pruning_details}.
We show detailed information about the pruning of VGG-19 on the CIFAR-10 dataset.
We count the PFGDF's convolution filter information in 16's layers.
The trimming occurrence is mainly concentrated in the back end of the network.
We guess that part of the reason is that  the number of convolution filters on the back of the network are much more
than on the front of the network
when the network was designed.
PFGDF compresses results show that these designs have a lot of redundant semantic information.
With the network deepening, the number of convolution filters at each layer shows an increasing trend.
The numbers retained are more than the number of front-end networks.
This means that more abstract semantic information is extracted with the deepening of the network, which is very important for image recognition.
We reveal the accuracy recovery details for each layer during the compression.

As shown in Fig.~\ref{fig:experiment_vgg19_retraining_acc}.
On CIFAR-10, the accuracy curves is showed for VGG-19 whne appropriate compression hyper-parameters are found in each layer.
The diagram shows that the convolutional layers at 3, 6, 10  are relatively slow in the early stage of recovery but the others layers after compression are very fast.
This shows that the small degree of front-end network compression
does not speed up the accuracy recovery of the model, and there is no relationship between them.
After the compression of almost all convolutional layers, the accuracy of the model recovers more than $ 90 \% $ at 80 epochs and is close to the original model performance.
Meanwhile, the process of restoration will be ended before the preset 160 epochs,
and then the pruning operation of the next layer is entered.
This also indicates that the structural information retained by PFGDF has better model performance recovery capabilities.

Further more, as shown in Table.~\ref{tab:Comparing_acceleration}, we also report the actual acceleration performance. Using MNN mobile terminal reasoning framework and compiling with armeabi-v7a instructions, we test the realistic acceleration performance on the Hisilicon Kirin970 CPU.
These experiments prove that PFGDF can effectively prune a lot of invalid information in the model and save the others with better characterization capabilities, which can efficiently restore the original performance of the model.The pruned model can achieve significant reasoning acceleration.

\begin{table}[]
	\caption{
		Realistic reasoning for acceleration performance. Using MNN reasoning framework, we test the experimental results on Huawei MATE 10, which is equipped with Hisilicon Kirin970 CPU.
}
	\vspace*{-1.0mm}
	\centering
	\noindent\resizebox{1.0\linewidth}{!}{
	\tabulinesep=0.8mm
	\renewcommand{\arraystretch}{1.6}
	\begin{tabular}{lllll}
\Xhline{0.1em}
		\textbf{Dataset}           & \textbf{Model} & \textbf{\begin{tabular}[c]{@{}l@{}}Baseline \\ time (MS)\end{tabular}} & \textbf{\begin{tabular}[c]{@{}l@{}}Pruned \\ time(MS)\end{tabular}} & \textbf{\begin{tabular}[c]{@{}l@{}}Realistic\\ Speed-up(\%)\end{tabular}} \\
\Xhline{0.1em}
		\multirow{4}{*}{CIFAR-10}  & VGG-16         & 36.582                                                                 & 5.953                                                               & 83.73\%                                                                   \\
		& VGG-19         & 44.73                                                                  & 8.982                                                               & 79.92\%                                                                   \\
		& ResNet-20      & 3.19                                                                   & 2.561                                                               & 19.72\%                                                                   \\
		& ResNet-32      & 5.214                                                                  & 4.703                                                               & 9.80\%                                                                    \\
\Xhline{0.1em}
		\multirow{2}{*}{CIFAR-100} & VGG-16         & 35.84                                                                  & 9.128                                                               & 74.53\%                                                                   \\
		& VGG-19         & 45.175                                                                 & 8.031                                                               & 82.22\%  \\
\Xhline{0.1em}
	\end{tabular}
	}
	\begin{tablenotes}
	\item  {
		{The result of the measurement is the average of 10 inferences on the CPU}}
	\end{tablenotes}

	\label{tab:Comparing_acceleration}
\end{table}

\section{Conclusion}

We propose a pruning filter method via gaussian distribution feature(PFGDF) for removing invalid and cumbersome filters in convolutional neural networks.
We extract  the distribution features of the trained model and analyze the changes in the convolution filter during the Gaussian distribution.
Based on the changing characteristics, we put forward the hypothesis and check the rationality.
PFGDF is an automatic iterative pruning method. During the pruning, it is always guaranteed 
that the compressed model has no performance loss. 
A grid search algorithm is adopted in the compression for hyper-parameter without human intervention.
This also means that fine-tuning after compression is not necessary for PFGDF. 
If the compressed model is fine-tuned, the performance might still continue to improve.
The extensive experiments demonstrate the effectiveness of PFGDF.

\bibliographystyle{IEEEtran}
\bibliography{IEEEabrv,conference_101719}

\begin{thebibliography}{10}
\providecommand{\url}[1]{#1}
\csname url@samestyle\endcsname
\providecommand{\newblock}{\relax}
\providecommand{\bibinfo}[2]{#2}
\providecommand{\BIBentrySTDinterwordspacing}{\spaceskip=0pt\relax}
\providecommand{\BIBentryALTinterwordstretchfactor}{4}
\providecommand{\BIBentryALTinterwordspacing}{\spaceskip=\fontdimen2\font plus
\BIBentryALTinterwordstretchfactor\fontdimen3\font minus
  \fontdimen4\font\relax}
\providecommand{\BIBforeignlanguage}[2]{{%
\expandafter\ifx\csname l@#1\endcsname\relax
\typeout{** WARNING: IEEEtran.bst: No hyphenation pattern has been}%
\typeout{** loaded for the language `#1'. Using the pattern for}%
\typeout{** the default language instead.}%
\else
\language=\csname l@#1\endcsname
\fi
#2}}
\providecommand{\BIBdecl}{\relax}
\BIBdecl

\bibitem{simonyan2014very}
K.~Simonyan and A.~Zisserman, ``Very deep convolutional networks for
  large-scale image recognition,'' \emph{arXiv preprint arXiv:1409.1556}, 2014.

\bibitem{wortsman2022model}
M.~Wortsman, G.~Ilharco, S.~Y. Gadre, R.~Roelofs, R.~Gontijo-Lopes, A.~S.
  Morcos, H.~Namkoong, A.~Farhadi, Y.~Carmon, S.~Kornblith \emph{et~al.},
  ``Model soups: averaging weights of multiple fine-tuned models improves
  accuracy without increasing inference time,'' \emph{arXiv preprint
  arXiv:2203.05482}, 2022.

\bibitem{sun2018face}
X.~Sun, P.~Wu, and S.~C. Hoi, ``Face detection using deep learning: An improved
  faster rcnn approach,'' \emph{Neurocomputing}, vol. 299, pp. 42--50, 2018.

\bibitem{mare2021realistic}
T.~Mare, G.~Duta, M.-I. Georgescu, A.~Sandru, B.~Alexe, M.~Popescu, and R.~T.
  Ionescu, ``A realistic approach to generate masked faces applied on two novel
  masked face recognition data sets,'' \emph{arXiv preprint arXiv:2109.01745},
  2021.

\bibitem{deshmukh2018yolo}
A.~Deshmukh and P.~Nair, ``Yolo: Unified, real-time object detection,'' 2018.

\bibitem{yang2021focal}
J.~Yang, C.~Li, P.~Zhang, X.~Dai, B.~Xiao, L.~Yuan, and J.~Gao, ``Focal
  self-attention for local-global interactions in vision transformers,''
  \emph{arXiv preprint arXiv:2107.00641}, 2021.

\bibitem{liu2021group}
L.~Liu, S.~Zhang, Z.~Kuang, A.~Zhou, J.-H. Xue, X.~Wang, Y.~Chen, W.~Yang,
  Q.~Liao, and W.~Zhang, ``Group fisher pruning for practical network
  compression,'' in \emph{International Conference on Machine Learning}.\hskip
  1em plus 0.5em minus 0.4em\relax PMLR, 2021, pp. 7021--7032.

\bibitem{he2019filter}
Y.~He, P.~Liu, Z.~Wang, Z.~Hu, and Y.~Yang, ``Filter pruning via geometric
  median for deep convolutional neural networks acceleration,'' in
  \emph{Proceedings of the IEEE Conference on Computer Vision and Pattern
  Recognition}, 2019, pp. 4340--4349.

\bibitem{he2017channel}
Y.~He, X.~Zhang, and J.~Sun, ``Channel pruning for accelerating very deep
  neural networks,'' in \emph{Proceedings of the IEEE International Conference
  on Computer Vision}, 2017, pp. 1389--1397.

\bibitem{diffenderfer2021multi}
J.~Diffenderfer and B.~Kailkhura, ``Multi-prize lottery ticket hypothesis:
  Finding accurate binary neural networks by pruning a randomly weighted
  network,'' \emph{arXiv preprint arXiv:2103.09377}, 2021.

\bibitem{sanh2020movement}
V.~Sanh, T.~Wolf, and A.~Rush, ``Movement pruning: Adaptive sparsity by
  fine-tuning,'' \emph{Advances in Neural Information Processing Systems},
  vol.~33, pp. 20\,378--20\,389, 2020.

\bibitem{zhu2017prune}
M.~Zhu and S.~Gupta, ``To prune, or not to prune: exploring the efficacy of
  pruning for model compression,'' \emph{arXiv preprint arXiv:1710.01878},
  2017.

\bibitem{han2015deep}
S.~Han, H.~Mao, and W.~J. Dally, ``Deep compression: Compressing deep neural
  networks with pruning, trained quantization and huffman coding,'' \emph{arXiv
  preprint arXiv:1510.00149}, 2015.

\bibitem{he2018soft}
Y.~He, G.~Kang, X.~Dong, Y.~Fu, and Y.~Yang, ``Soft filter pruning for
  accelerating deep convolutional neural networks,'' \emph{arXiv preprint
  arXiv:1808.06866}, 2018.

\bibitem{li2016pruning}
\BIBentryALTinterwordspacing
H.~Li, A.~Kadav, I.~Durdanovic, H.~Samet, and H.~P. Graf, ``Pruning filters for
  efficient convnets,'' \emph{CoRR}, vol. abs/1608.08710, 2016. [Online].
  Available: \url{http://arxiv.org/abs/1608.08710}
\BIBentrySTDinterwordspacing

\bibitem{glorot2018understanding}
X.~Glorot and Y.~Bengio, ``Understanding the difficulty of training deep
  feedforward neural networks. 2010,'' in \emph{International Conference on
  Artificial Intelligence and Statistics}, 2018.

\bibitem{he2018delving}
K.~He, X.~Zhang, S.~Ren, and J.~Sun, ``Delving deep into rectifiers: surpassing
  human-level performance on imagenet classification. arxiv e-prints 2015,''
  2018.

\bibitem{Liu_2017_ICCV}
Z.~Liu, J.~Li, Z.~Shen, G.~Huang, S.~Yan, and C.~Zhang, ``Learning efficient
  convolutional networks through network slimming,'' in \emph{The IEEE
  International Conference on Computer Vision (ICCV)}, Oct 2017.

\bibitem{louizos2017learning}
C.~Louizos, M.~Welling, and D.~P. Kingma, ``Learning sparse neural networks
  through $ l\_0 $ regularization,'' \emph{ICLR}, 2018.

\bibitem{hanson1989comparing}
S.~J. Hanson and L.~Y. Pratt, ``Comparing biases for minimal network
  construction with back-propagation,'' in \emph{Advances in neural information
  processing systems}, 1989, pp. 177--185.

\bibitem{lecun1990optimal}
Y.~LeCun, J.~S. Denker, and S.~A. Solla, ``Optimal brain damage,'' in
  \emph{Advances in neural information processing systems}, 1990, pp. 598--605.

\bibitem{hassibi1993second}
B.~Hassibi and D.~G. Stork, ``Second order derivatives for network pruning:
  Optimal brain surgeon,'' in \emph{Advances in neural information processing
  systems}, 1993, pp. 164--171.

\bibitem{han2015learning}
S.~Han, J.~Pool, J.~Tran, and W.~Dally, ``Learning both weights and connections
  for efficient neural network,'' in \emph{Advances in neural information
  processing systems}, 2015, pp. 1135--1143.

\bibitem{sehwag2020hydra}
V.~Sehwag, S.~Wang, P.~Mittal, and S.~Jana, ``Hydra: Pruning adversarially
  robust neural networks,'' 2020.

\bibitem{tang2021manifold}
Y.~Tang, Y.~Wang, Y.~Xu, Y.~Deng, C.~Xu, D.~Tao, and C.~Xu, ``Manifold
  regularized dynamic network pruning,'' in \emph{Proceedings of the IEEE/CVF
  Conference on Computer Vision and Pattern Recognition}, 2021, pp. 5018--5028.

\bibitem{han2016eie}
S.~Han, X.~Liu, H.~Mao, J.~Pu, A.~Pedram, M.~A. Horowitz, and W.~J. Dally,
  ``Eie: efficient inference engine on compressed deep neural network,'' in
  \emph{2016 ACM/IEEE 43rd Annual International Symposium on Computer
  Architecture (ISCA)}.\hskip 1em plus 0.5em minus 0.4em\relax IEEE, 2016, pp.
  243--254.

\bibitem{DBLP:journals/corr/abs-1803-03635}
\BIBentryALTinterwordspacing
J.~Frankle and M.~Carbin, ``The lottery ticket hypothesis: Training pruned
  neural networks,'' \emph{CoRR}, vol. abs/1803.03635, 2018. [Online].
  Available: \url{http://arxiv.org/abs/1803.03635}
\BIBentrySTDinterwordspacing

\bibitem{liu2018rethinking}
Z.~Liu, M.~Sun, T.~Zhou, G.~Huang, and T.~Darrell, ``Rethinking the value of
  network pruning,'' in \emph{ICLR}, 2019.

\bibitem{krizhevsky2012imagenet}
A.~Krizhevsky, I.~Sutskever, and G.~E. Hinton, ``Imagenet classification with
  deep convolutional neural networks,'' in \emph{Advances in neural information
  processing systems}, 2012, pp. 1097--1105.

\bibitem{he2016deep}
K.~He, X.~Zhang, S.~Ren, and J.~Sun, ``Deep residual learning for image
  recognition,'' in \emph{Proceedings of the IEEE conference on computer vision
  and pattern recognition}, 2016, pp. 770--778.

\bibitem{szegedy2015going}
C.~Szegedy, W.~Liu, Y.~Jia, P.~Sermanet, S.~Reed, D.~Anguelov, D.~Erhan,
  V.~Vanhoucke, and A.~Rabinovich, ``Going deeper with convolutions,'' in
  \emph{Proceedings of the IEEE conference on computer vision and pattern
  recognition}, 2015, pp. 1--9.

\bibitem{lin2013network}
M.~Lin, Q.~Chen, and S.~Yan, ``Network in network,'' \emph{arXiv preprint
  arXiv:1312.4400}, 2013.

\bibitem{kim2020neuron}
W.~Kim, S.~Kim, M.~Park, and G.~Jeon, ``Neuron merging: Compensating for pruned
  neurons,'' 2020.

\bibitem{zhao2019variational}
C.~Zhao, B.~Ni, J.~Zhang, Q.~Zhao, W.~Zhang, and Q.~Tian, ``Variational
  convolutional neural network pruning,'' in \emph{Proceedings of the IEEE
  Conference on Computer Vision and Pattern Recognition}, 2019, pp. 2780--2789.

\bibitem{lin2019towards}
S.~Lin, R.~Ji, C.~Yan, B.~Zhang, L.~Cao, Q.~Ye, F.~Huang, and D.~Doermann,
  ``Towards optimal structured cnn pruning via generative adversarial
  learning,'' in \emph{Proceedings of the IEEE Conference on Computer Vision
  and Pattern Recognition}, 2019, pp. 2790--2799.

\bibitem{huang2018data}
Z.~Huang and N.~Wang, ``Data-driven sparse structure selection for deep neural
  networks,'' in \emph{Proceedings of the European Conference on Computer
  Vision (ECCV)}, 2018, pp. 304--320.

\bibitem{xu2018globally}
K.~Xu, X.~Wang, Q.~Jia, J.~An, and D.~Wang, ``Globally soft filter pruning for
  efficient convolutional neural networks,'' 2018.

\bibitem{wang2019eigendamage}
C.~Wang, R.~Grosse, S.~Fidler, and G.~Zhang, ``Eigendamage: Structured pruning
  in the kronecker-factored eigenbasis,'' \emph{arXiv preprint
  arXiv:1905.05934}, 2019.

\bibitem{zhuang2020neuron}
T.~Zhuang, Z.~Zhang, Y.~Huang, X.~Zeng, K.~Shuang, and X.~Li, ``Neuron-level
  structured pruning using polarization regularizer,'' \emph{Advances in Neural
  Information Processing Systems}, vol.~33, 2020.

\bibitem{szegedy2016rethinking}
C.~Szegedy, V.~Vanhoucke, S.~Ioffe, J.~Shlens, and Z.~Wojna, ``Rethinking the
  inception architecture for computer vision,'' in \emph{Proceedings of the
  IEEE conference on computer vision and pattern recognition}, 2016, pp.
  2818--2826.

\bibitem{NeuraArtisticStyle2015l}
L.~A. Gatys, A.~S. Ecker, and M.~Bethge, ``A neural algorithm of artistic
  style,'' 2015.

\end{thebibliography}

\begin{appendices} 	

\begin{figure*}
	\begin{center}
		\includegraphics[width=1\linewidth]{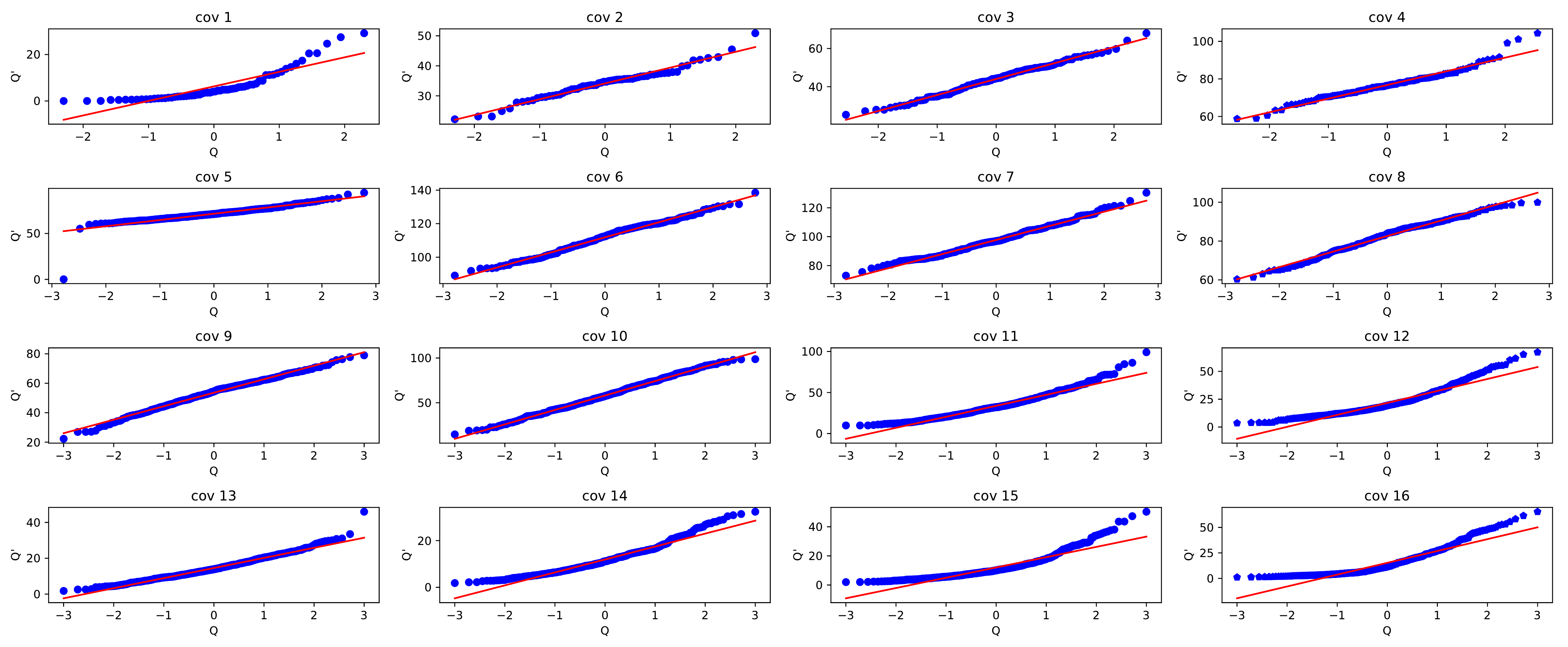}\\		
	\end{center}
	\caption{
			Illustration of hypothetical rationality.
			On CIFAR100 dataset, all the convolutional layers of VGG-19 were checked for the rationality of the Gaussian distribution on the QQ diagram. 
			The points corresponding to $ Q-Q'$ are approximately distributed near the straight line.
	}
	\label{fig:vgg_gaussian_check_cifar100}
\end{figure*}	

\begin{figure}[t]
	\begin{center}
		\subfigure[initialization]{
			\begin{minipage}[t]{0.5\linewidth}
				\centering
				\includegraphics[width=1.5in]{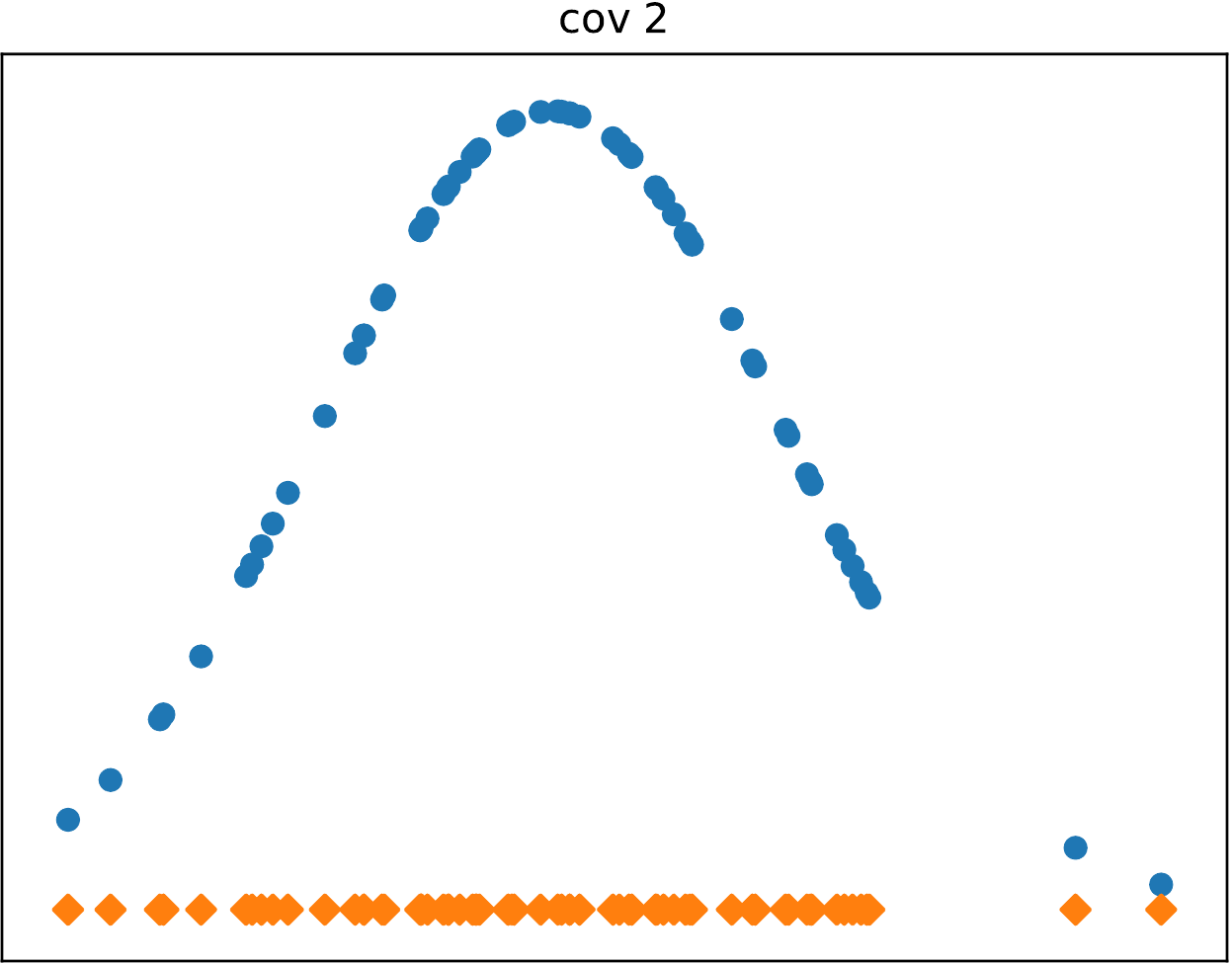}
			\end{minipage}%
		}%
		\subfigure[16 epochs]{
			\begin{minipage}[t]{0.5\linewidth}
				\centering
				\includegraphics[width=1.5in]{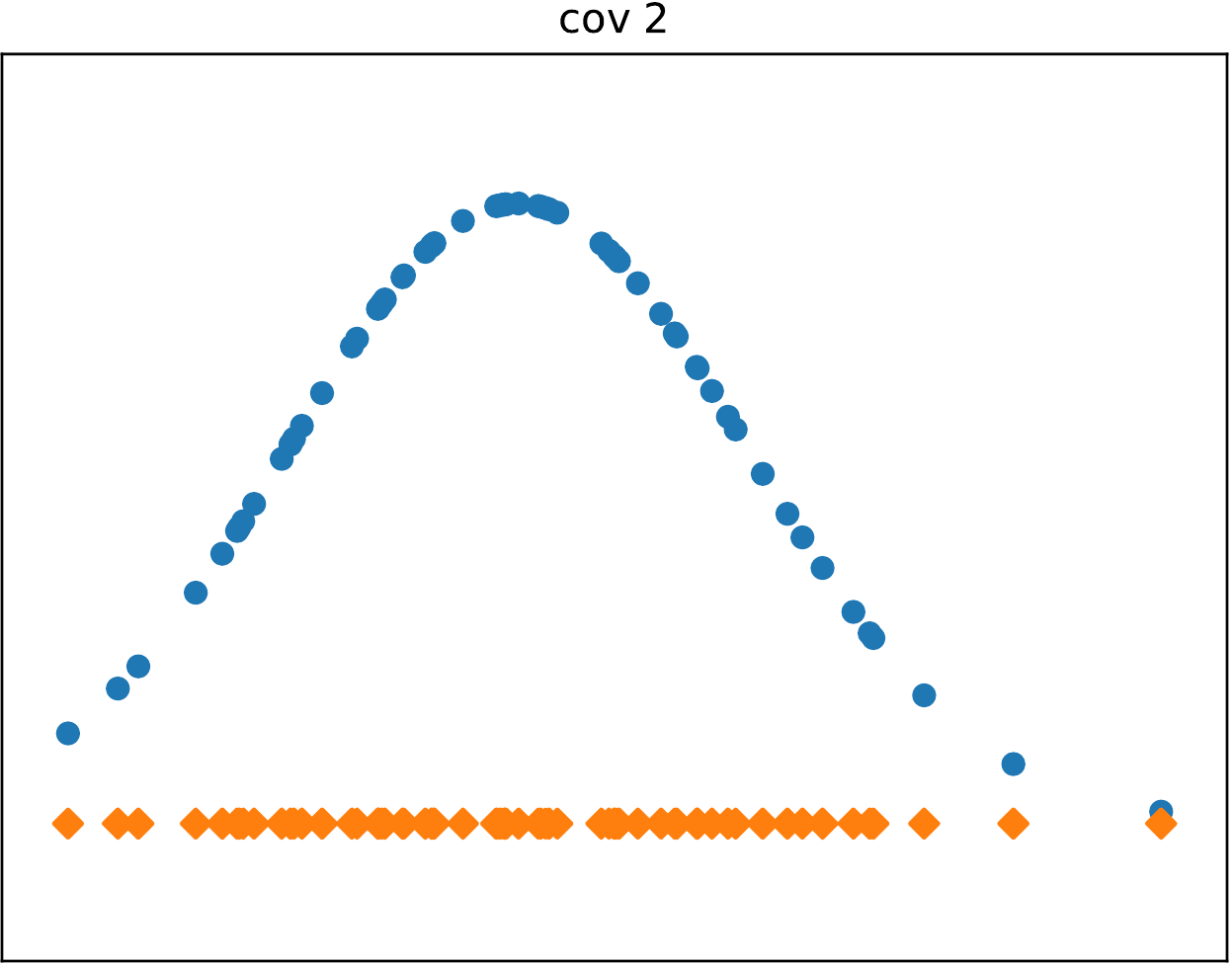}
			\end{minipage}%
		}%
		
		\subfigure[32 epochs]{
			\begin{minipage}[t]{0.5\linewidth}
				\centering
				\includegraphics[width=1.5in]{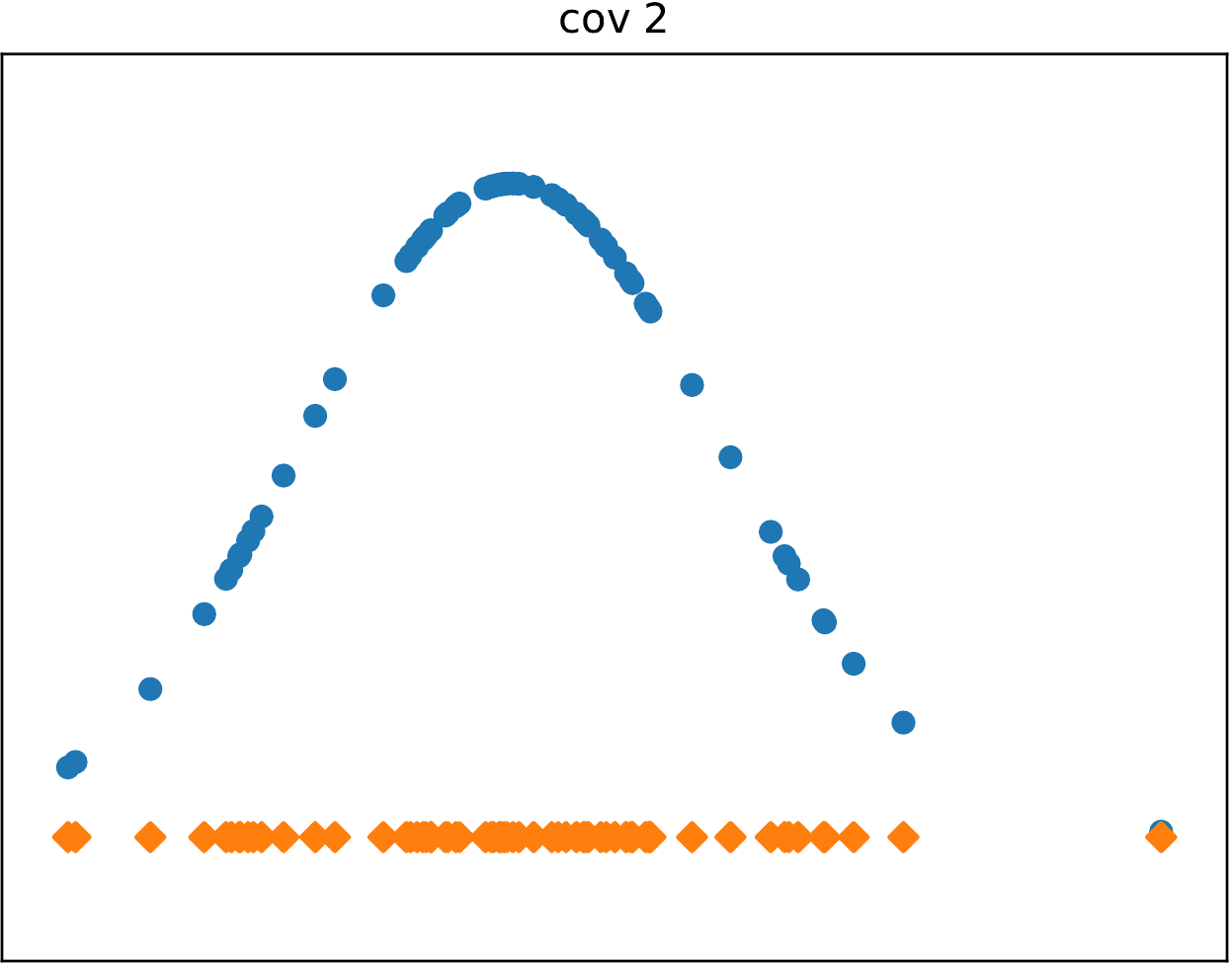}
			\end{minipage}
		}%
		\subfigure[160 epochs]{
			\begin{minipage}[t]{0.5\linewidth}
				\centering
				\includegraphics[width=1.5in]{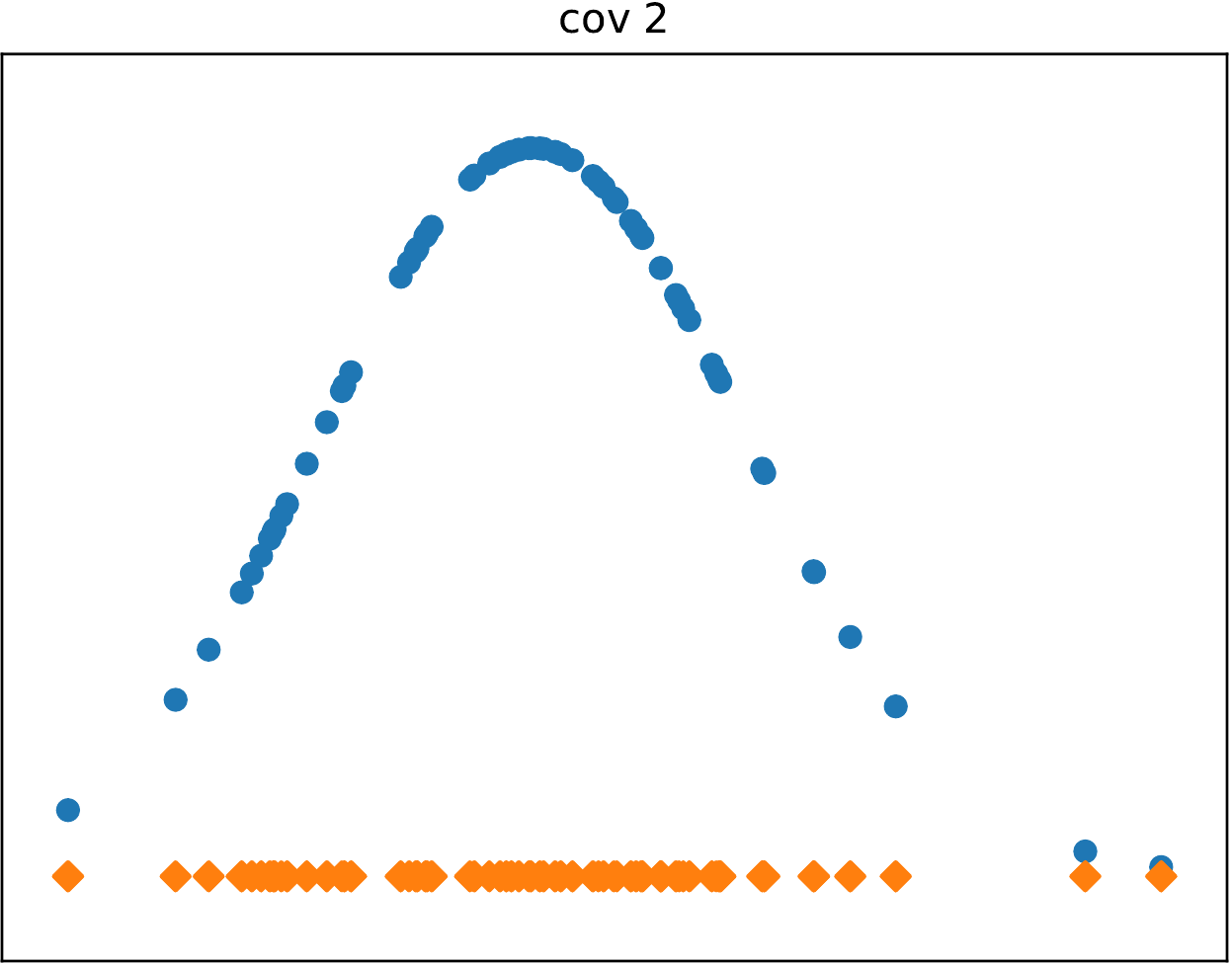}
			\end{minipage}
		}%
	\end{center}
	\caption{
		Illustrating the convolution filter $ f (i, j) $ 's Gaussian distribution during training.
		In CIFAR-100, the distribution of the second-level convolution filter of VGG-16 during the training processes.
		Blue: Gaussian distribution characteristics changing. 
		Orange: Density changing.	
	}
	\label{fig:show_gaussian_cifar100_vgg16}
\end{figure}	
%

	\section{Supplement of Hypothesis 1: Distribution Hypothesis} 
\begin{figure*}
	\begin{center}
		\includegraphics[width=1\linewidth]{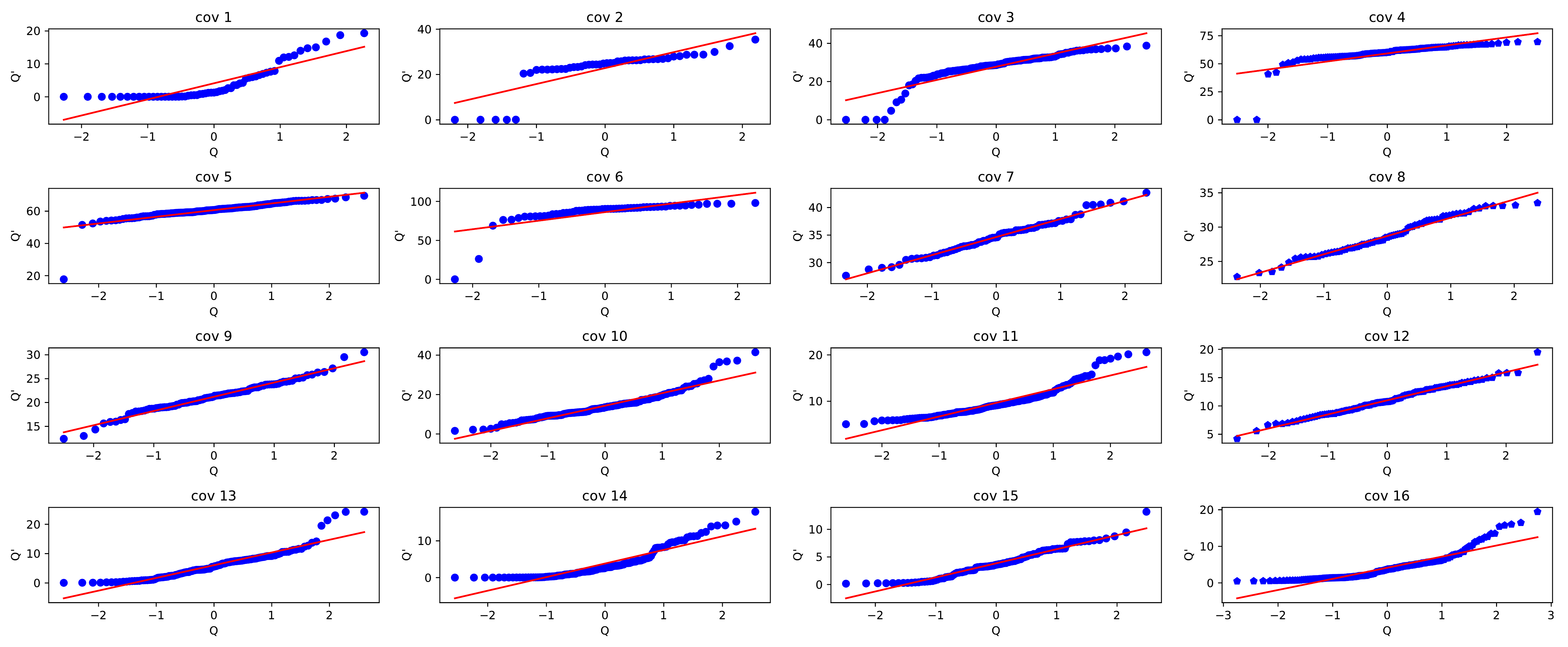}\\		
	\end{center}
	\caption{
			Illustration of hypothetical rationality.
			On CIFAR10 dataset, all the convolutional layers of VGG-19 were checked for the rationality of the Gaussian distribution on the QQ diagram. 
			The points corresponding to $ Q-Q'$ are approximately distributed near the straight line.
	}
	\label{fig:vgg_gaussian_check_cifar10}
\end{figure*}

%
	In this part, we will supplement the rationality of \textbf {Hypothesis 1: Distribution Hypothesis}. We have observed the distribution characteristics of the model after convergence in 
	different datasets to 
	explain the rationality of hypothesis 1.
	In Fig.~\ref{fig:vgg_gaussian_check}, we show the Gaussian rationality information of VGG-19 after training convergence on the ImageNet LSVRC 2012 dataset. In this supplementary section, we will supplement to show whether it has the same performance under other datasets. 
	Fig.~\ref{fig:vgg_gaussian_check_cifar100} shows the Gaussian rationality information of VGG-19 after CIFAR-100 dataset training convergence.
	The Gaussian rationality of VGG-19 training convergence on the CIFAR-10 dataset will be shown in Fig. ~\ref{fig:vgg_gaussian_check_cifar10}.
	By comparing the Fig. ~\ref{fig:vgg_gaussian_check}, the Fig.~\ref{fig:vgg_gaussian_check_cifar100} and the Fig.~\ref{fig:vgg_gaussian_check_cifar10},
	We find that the distribution of each layer of convolution filters on the ImageNet LSVRC 2012 dataset is almost near the QQ diagram. Fig. ~\ref{fig:vgg_gaussian_check_cifar100} performs worse and 
	Fig.~\ref{fig:vgg_gaussian_check_cifar10} performs the worst.We think this phenomenon may be related to the complexity of the training dataset and the number of convolution filters in each layer.  
	The higher the complexity of the training dataset, there are more convolution filters that need to actively participate in the process of extracting feature information.
	If the complexity of dataset is lower, there are less convolution filters of each layer that actually participates in the extraction of data feature information.
	Therefore, the distribution of the QQ diagram will be affected by the convolution filters that have not learned any information or have not learned  information that contributes to network reasoning.
	
	In general, most of the convolution filters in convolution neural networks are distributed around QQ diagram.
	Although we did not conduct further reasonable confidence test, we believe, from the experimental phenomenon, 
	that this hypothesis can be used to guide model compression pruning process, so as to achieve model compression and achieve the purposes of model optimization.

\begin{figure}[t]
	\begin{center}
		\subfigure[initialization]{
			\begin{minipage}[t]{0.5\linewidth}
				\centering
				\includegraphics[width=1.5in]{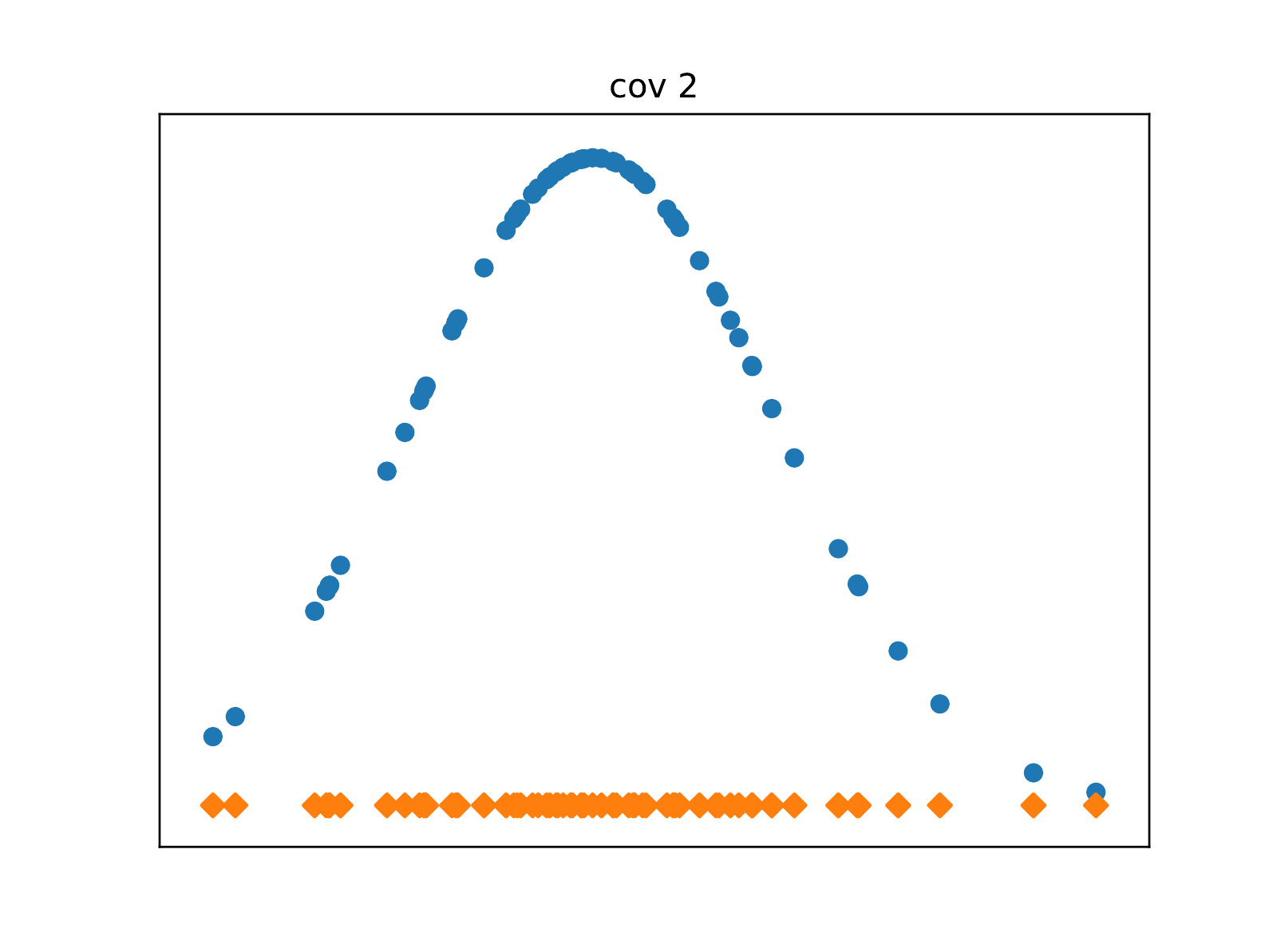}
			\end{minipage}%
		}%
		\subfigure[16 epochs]{
			\begin{minipage}[t]{0.5\linewidth}
				\centering
				\includegraphics[width=1.5in]{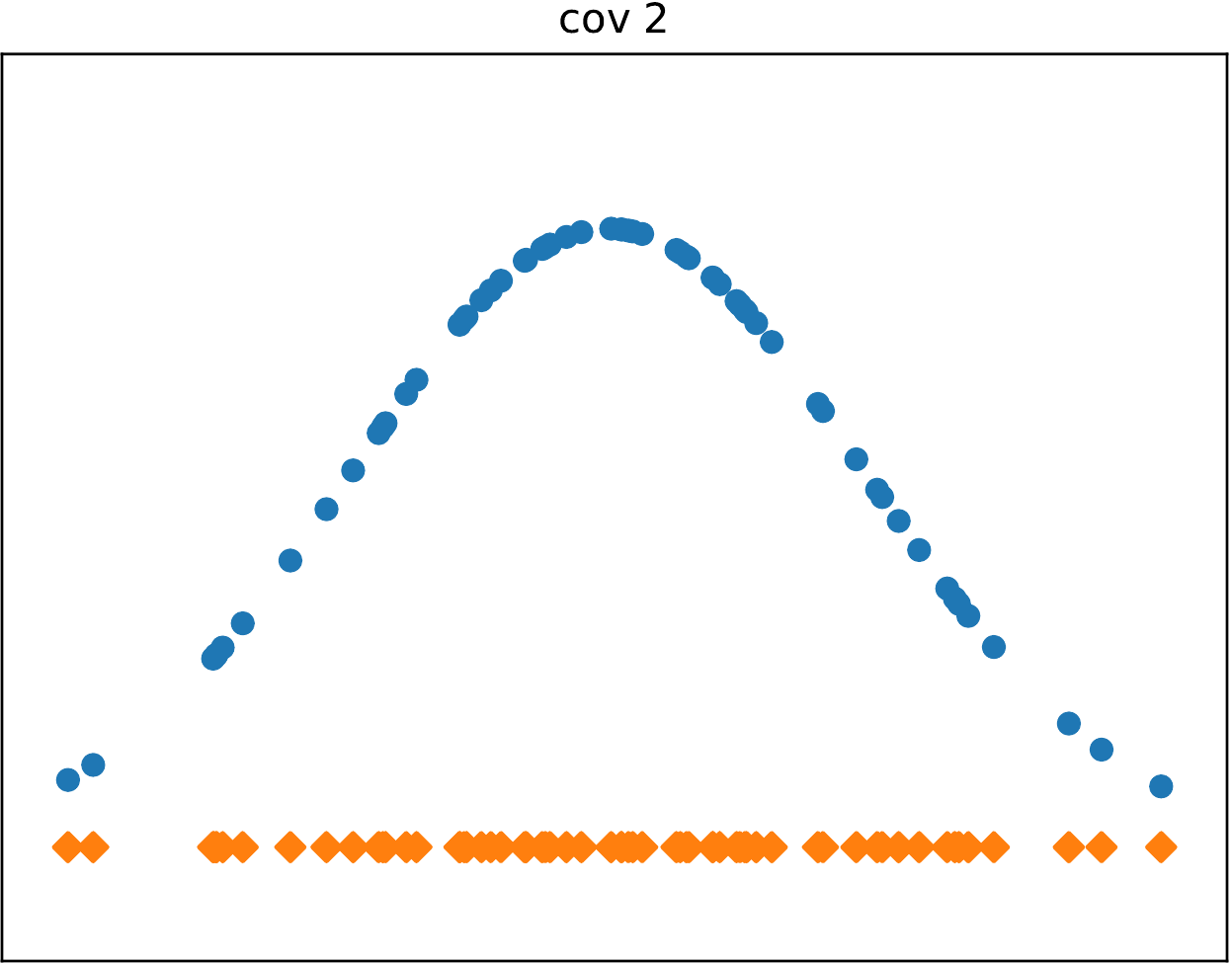}
			\end{minipage}%
		}%
		
		\subfigure[32 epochs]{
			\begin{minipage}[t]{0.5\linewidth}
				\centering
				\includegraphics[width=1.5in]{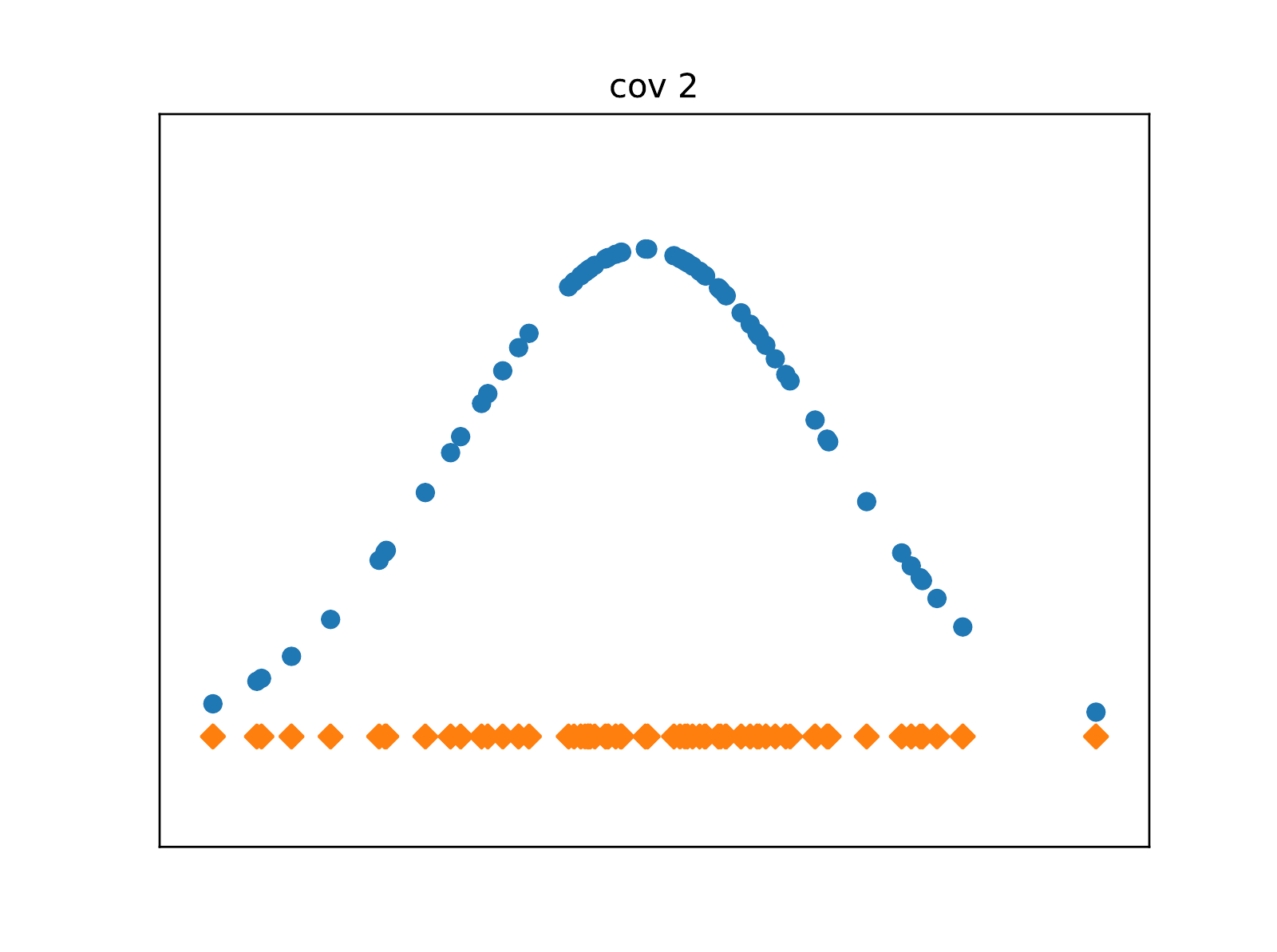}
			\end{minipage}
		}%
		\subfigure[160 epochs]{
			\begin{minipage}[t]{0.5\linewidth}
				\centering
				\includegraphics[width=1.5in]{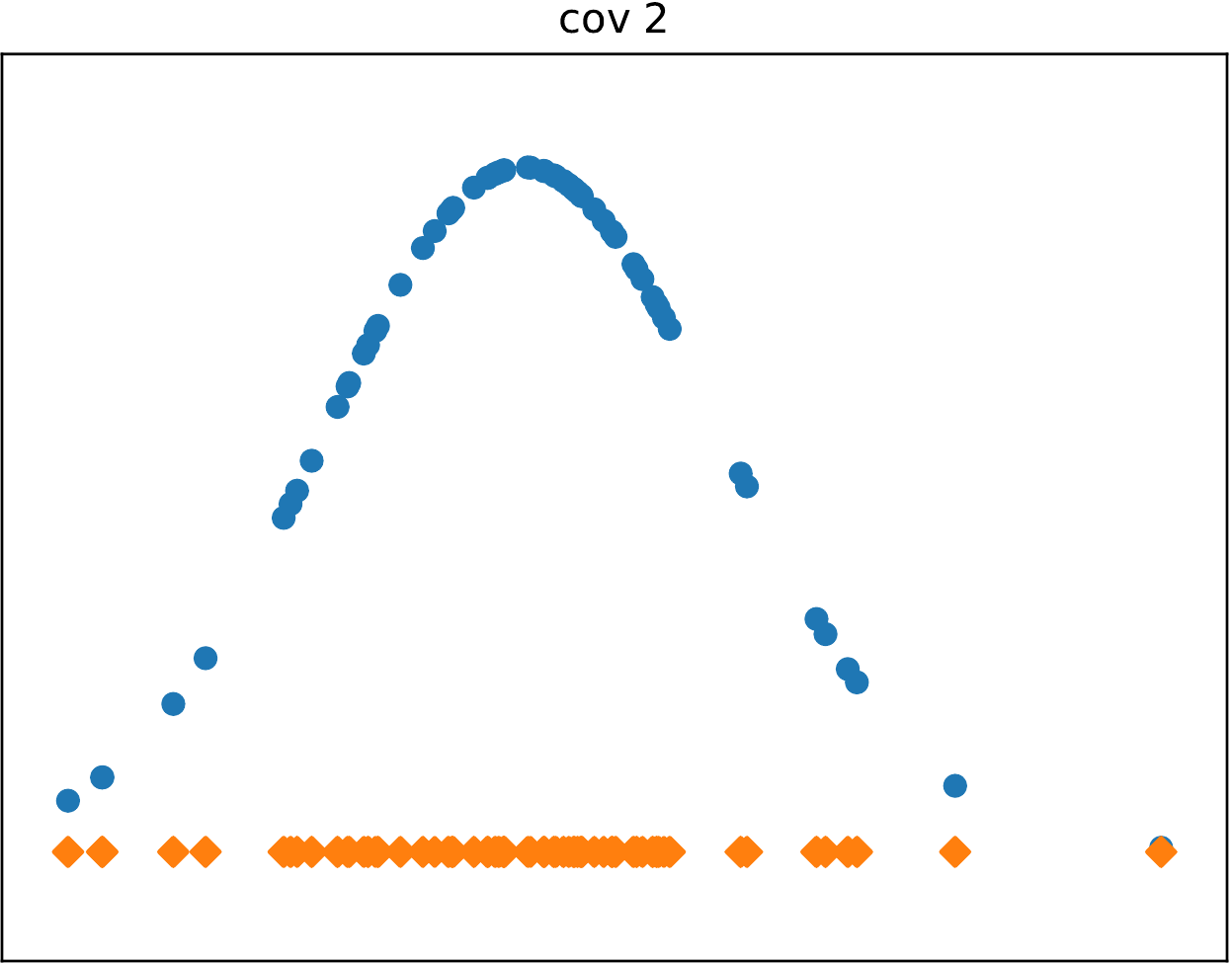}
			\end{minipage}
		}%
	\end{center}
	\caption{
		Illustrating the convolution filter $ f (i, j) $ 's Gaussian distribution during training.
		In CIFAR-100, the distribution of the second-level convolution filter of VGG-19 during the training processes.
		Blue: Gaussian distribution characteristics changing. 
		Orange: Density changing.	
	}
	\label{fig:show_gaussian_cifar100_vgg19}
\end{figure}	
%
\section{Supplement of Hypothesis 2: Convergence Hypothesis} 
	In this section, we will supplement \textbf{Hypothesis 2: Convergence Hypothesis} in the paper.
	We observe the convergence of different network under the 10 classification (CIFAR-10), 100 classification (CIFAR-100) and 1000 classification (ImageNet LSVRC 2012) dataset.
	They almost have the same convergence characteristics.
	Here, we supplement the performance of different VGG network structures under the CIFAR-100 dataset.
	Fig.~\ref{fig:show_gaussian_cifar100_vgg16} shows that the distribution of the second-level convolution filters of VGG-16 during the training processes.
	   The distribution of the second-level convolution filters of VGG-19 during the training processes is showed by Fig.~\ref{fig:show_gaussian_cifar100_vgg19}.
	   Fig.~\ref{fig:show_gaussian_cifar100_vgg16} and Fig.~\ref{fig:show_gaussian_cifar100_vgg19} show that $\mathcal {F} _ {i} $ appeares to
		 converge to a certain center of the Gaussian distribution during training process.
	In fact, under the same dataset, the model almost presents this convergence feature during the training process.  They are the same in different networks.
	The distributional features and the normality hold are for different initializations, different learning rate, activation functions, optimizer and etc. 
	From the perspective of different parameter information, the appropriate hyperparameters are to enable the model to converge at the end of training and to learn the sample information of the training dataset.
	From the perspective of the model training results, when the same model is converged, the weight information of the convolution filters of each layer maybe converge within a certain range, and different hyperparameters almosts hope that the convolution filters of the model can find such weight information.
	At this point, we find some verification in traditional image processing. 
	Manual design of convolution kernel templates almost obey certain rules.
	For example, when manually designing low-pass convolution kernel templates, they are usually set to the norm sum of L1-norm empirically.
	
	It should be noted that the phenomenon that the convolution filter converges toward the Gaussian center does not mean that the convolution filter is similar when extracting information.
	Because These phenomena only indicate convergence information and do not have the similarity feature of the convolution kernel or the similarity information of the feature map.
	 In recent works, the pruning method of geometric median~\cite{he2019filter} indicates that these convolution filters are similar by calculating their the Euclidean distance, 
	 and similar convolution filters can be cut off.
	 However, our method only shows that these convolution filters have convergence phenomena on L1-norm, while convolution filters that cannot converge may be redundant and can be tailored safely.
	In summary, the rationality of \textbf{Hypothesis 2: Convergence Hypothesis} is explained by supplementing the characteristics of different VGG network structures in different datasets, 
	and this hypothesis can be used to guide model compression pruning process, so as to achieve model compression and achieve the purposes of model optimization.
\end{appendices}

\end{document}